\pdfoutput=1

\documentclass[11pt]{article}

\usepackage{emnlp2021}

\usepackage{times}
\usepackage{latexsym}

\usepackage[T1]{fontenc}

\usepackage[utf8]{inputenc}

\usepackage{microtype}

\usepackage{graphicx}
\usepackage{amsmath}
\usepackage{amssymb}
\usepackage{booktabs}
\usepackage{algorithm}
\usepackage{algorithmicx}
\usepackage{stmaryrd}
\usepackage[noend]{algpseudocode}
\usepackage{capt-of}

\usepackage{soul}
\usepackage{multirow}
\usepackage{tabularx}

\newcommand{\eg}{\textit{e}.\textit{g}.}
\newcommand{\ie}{\textit{i}.\textit{e}.}
\newcommand{\bgray}[1]{{\color{gray}\textbf{#1}}}

\DeclareMathOperator{\knn}{\operatorname{KNN}}

\newcommand{\down}[1]{{\color{red}$\downarrow$ #1}}

\makeatletter
\g@addto@macro{\endtabular}{\rowfont{}}
\makeatother
\newcommand{\rowfonttype}{}
\newcommand{\rowfont}[1]{
   \gdef\rowfonttype{#1}#1%
}
\newcolumntype{L}{>{\rowfonttype}l}

\title{Cross-Modal Retrieval Augmentation for Multi-Modal Classification}

\author{Shir Gur$^\dagger$, 
Natalia Neverova$^\ddagger$, Chris Stauffer$^\ddagger$, 
Ser-Nam Lim$^\ddagger$, Douwe Kiela$^\ddagger$, Austin Reiter$^\ddagger$\\
$^\dagger$Tel Aviv University; $^\ddagger$Facebook AI\\
\texttt{shir.gur@cs.tau.ac.il}\\
\texttt{\{nneverova,cstauffer,sernamlim,dkiela,areiter\}@fb.com}
}
\begin{document}
\maketitle

\begin{abstract}
Recent advances in using retrieval components over external knowledge sources have shown impressive results for a variety of downstream tasks in natural language processing. Here, we explore the use of unstructured external knowledge sources of images and their corresponding captions for improving visual question answering (VQA). First, we train a novel alignment model for embedding images and captions in the same space, 
which achieves substantial improvement in performance on image-caption retrieval w.r.t. similar methods.
Second, we show that retrieval-augmented multi-modal transformers using the trained alignment model
improve results on VQA over strong baselines.
We further conduct extensive experiments to establish the promise of this approach, and examine novel applications for inference time such as hot-swapping indices.
\end{abstract}

\section{Introduction}

Neural networks augmented with non-parametric retrieval components have recently shown impressive results in NLP \cite{khandelwal2019generalization,guu2020realm,lewis2020retrieval,izacard2020leveraging}. 
In this work, we introduce a novel image-caption alignment model architecture and utilize it in various retrieval-augmented multi-modal transformer models, achieving substantial improvement over strong baselines.

Retrieval components are promising because they allow for easy revision and expansion of their memory, as compared to their parametric, pre-training counterparts.  
They provide more interpretability, as well as direct factual consistency with trusted knowledge sources. In the multi-modal setting, retrieval augmentation allows for leveraging the strengths of text-based models---as evidenced by the strong performance of BERT-based models in vision-and-language \cite{lu2019vilbert,li2019visualbert,kiela2019supervised}---via cross-modal translation from images to text.
Being able to seamlessly ``hot swap'' knowledge sources without the need for re-training the model affords a unique scalability not typically seen in the traditional deep learning literature. Nearest neighbor methods are known to be strong baselines in the vision and language domain~\cite{devlin2015exploring}.

We introduce a simple, yet effective, novel cross-modal alignment architecture called DXR~(Dense X-modal Retriever). DXR achieves a substantial increase in performance on both COCO~\cite{chen2015microsoft} and Flickr30k~\cite{young2014image} image-caption retrieval, {with respect to similar methods}. 
We subsequently use DXR to augment several multi-modal transformer architectures with a retrieval component. 
We show that retrieval augmentation yields impressive results for a variety of well-known multi-modal transformer architectures, ranging from VisualBERT~\cite{li2019visualbert} and ViLBERT~\cite{lu2019vilbert}---which use bounding-box features---to Movie+MCAN~\cite{nguyen2020revisiting}---which uses grid features. We name our overall method XTRA, for X-modal Transformer Retrieval Augmentation. Specifically, our contributions are as follows:
\begin{itemize}
    \item We introduce a novel image-caption retrieval architecture, DXR, that achieves a substantial increase in performance on COCO and Flickr30k, with respect to similar methods.
    \item We introduce a new retrieval-augmented multi-modal transformer architecture, XTRA, that results in noticeable improvements in VQA over the corresponding baselines, and is independent of the use of detection inputs. To our knowledge, this is the first work to showcase the promise of hybrid parametric and non-parametric models for the vision and language domain.
    \item We conduct extensive experiments to shed light on this novel approach. We explore different datasets for training the alignment model, as well as the effect of in-domain versus out-of-domain retrieval indices, the index size and inference time applications. 
    Our experiments show that the proposed method also preforms better than pre-training techniques, for the tested baselines.
\end{itemize}

\section{Related Work}
\label{sec:related}
\paragraph{Cross-Modal Retrieval}
Prior work in cross-modal retrieval can be divided into two primary categories: (i) methods that use grid-features and/or vector representations of the embedding space, and (ii) methods that use detection features, sequence representations, or share information between the two modalities for computing the similarity metric.

The first category consists of methods such as RRF~\cite{liu2017learning} and DPC~\cite{zheng2017dual} which use two network branches, for image and text. CMPM~\cite{zhang2018deep} introduced a Bi-directional LSTM to learn image and text embeddings. The most relevant work in this category is VSE++~\cite{faghri2017vse++}, which focuses on hard negative mining and ranking loss. Recently, two methods that uses substantial amount of data were proposed, CLIP~\cite{radford2021learning} which uses 0.4 Billion image-text pairs, and ALIGN~\cite{jia2021scaling} which uses 1.8 Billion noisy image-text pairs. Both methods use a dual encoder that produced and embedding vector for each modality. For fair comparison, and the lack of access to such data, we show performance against methods who use the common training setting on COCO and Flickr30K datasets.

The second category generally exploits the use of detection features, which enforces an additional complexity. Methods such as TERN~\cite{messina2020transformer}, TERAN~\cite{messina2020fine}, SAEM~\cite{wu2019learning} and MMCA~\cite{wei2020multi}, use transformer modules to obtain modality-specific embeddings. TERAN, as well as SCAN~\cite{lee2018stacked}, utilize sequence similarities. SCO~\cite{huang2018learning} and VSRN~\cite{li2019visual} learn, in addition to image-text alignment, to generate the caption from the image embedding. MMCA, as well as CAMP~\cite{wang2019camp}, fuses image and text information to obtain the final embeddings. VisualSparta~\cite{lu2021visualsparta} uses fragment-level interaction to compute similarity scores.
Other methods, such as Unicoder-VL~\cite{li2020unicoder}, Oscar~\cite{li2020oscar} and UNITER~\cite{chen2020uniter} are trained for multi-modal alignment as a pre-training task. While these models perform well, they suffer from high computational complexity as we discuss in Sec.~\ref{sec:timecop}.

\paragraph{External Knowledge Source Methods}
The use of an external knowledge source (KS) has gained much attention in the field of natural language processing (NLP), such as the work of~\citet{verga2020facts}. Our work is inspired by that of \citet{lewis2020retrieval}, which introduced RAG, a generic approach for a variety of downstream NLP tasks, that uses a learned retriever (DPR by~\citet{karpukhin2020dense}) to augment the inputs by marginalizing across several retrieved phrases retrieved from Wikipedia.
In the multi-modal domain, previous efforts have focused on building different types of KS, such as the work of~\citet{zhu2014reasoning,chen2013neil,divvala2014learning,sadeghi2015viske} and~\citet{zhu2015building}, which use web information for the construction of the KS. 
Methods that use an external KS for a downstream task use a structured KS, such as the work of~\citet{narasimhan2018out,narasimhan2018straight,wang2015explicit,wang2018fvqa} and \citet{zhu2017knowledge}. \citet{zhu2017knowledge} introduced an iterative method for VQA tasks. \citet{marino2019ok} introduced OK-VQA, a novel VQA dataset that requires the use of an external KS. \citet{fan2020augmenting} applied a KS to multi-modal dialogue.
In our work, we focus on a more natural KS, such as images and captions, which better reflect the data generated in newspapers and social media.

\begin{figure*}[t]
    \centering
    \begin{tabular}{@{}c@{~~~}c@{~~}c}
    \includegraphics[height=68px]{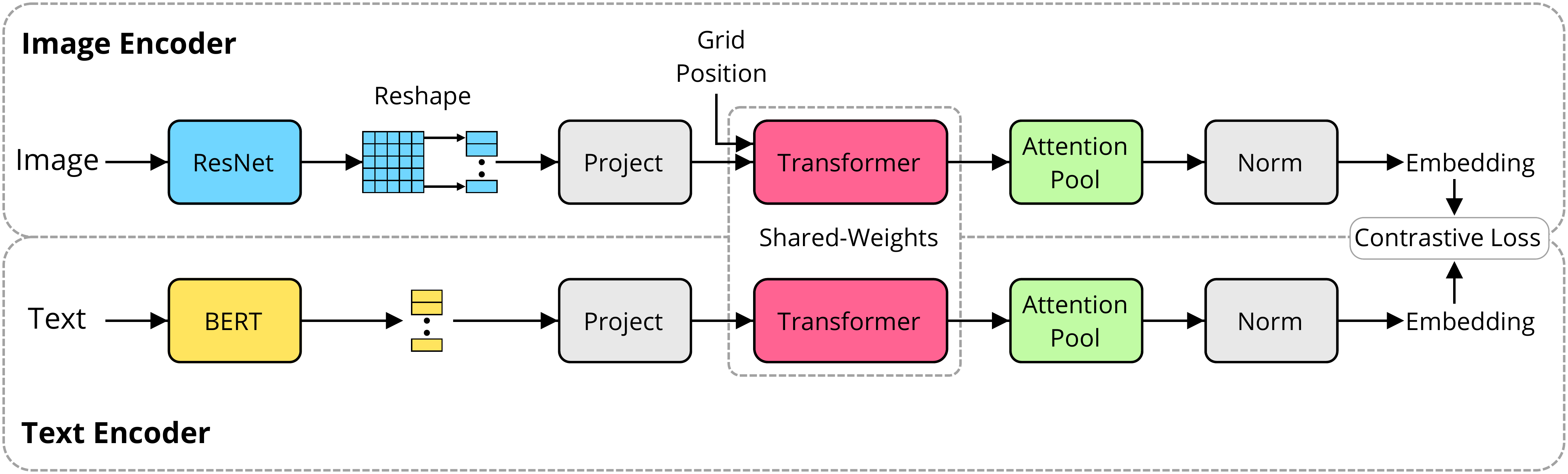}&
    \raisebox{15px}{\includegraphics[height=50px]{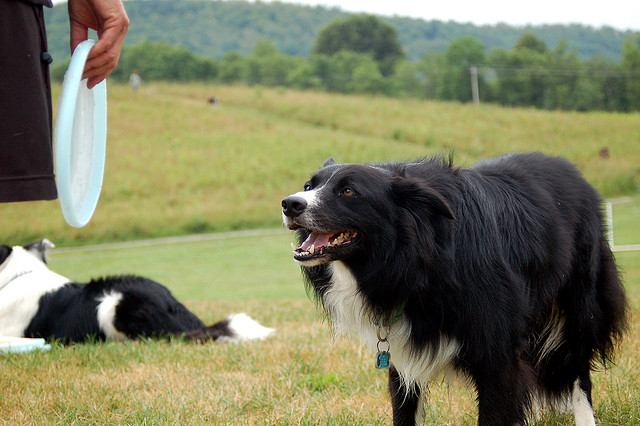}}&
    \raisebox{3px}{
    \begin{minipage}[b]{0.3\textwidth}
    \fontsize{7}{6} \selectfont 
    A black and white dog lays next to a frisbee\\
    
    {\color{blue}The black and white dog stands near\\ a person holding a frisbee}\\
    
    A black and white dog with a frisbee by its feet\\
    
    A black and white dog sitting next to a frisbee\\
    
    {\color{blue}A dog looking at a man holding a frisbee with\\ another dog laying down}
    \end{minipage}}\\
    (a) & \multicolumn{2}{c}{(b)}
    \end{tabular}
    \caption{(a) Cross-modal alignment architecture. We use a pre-trained ResNet-152 and BERT as feature extractors with an in-batch hinge loss. (b) Sample query image and retrieved captions from the COCO dataset. Ground truth captions are colored in blue (best viewed in color).}
    \label{fig:cross-modal}
\end{figure*}
\paragraph{Multi-modal Classification}
In this work, we investigate the potential advantages of using an external KS for the popular and challenging VQA domain, a multi-modal classification task. Current methods for VQA use pre-training on different datasets in order to gain better performance. 
In our experiments, we show performance for three different methods, (i) VisualBERT~\cite{li2019visualbert}, which is based on the BERT model by \citet{devlin2018bert}, (ii) ViLBERT~\cite{lu2019vilbert}, which fuses text and image modalities using co-attentional transformer layers, and (iii) MoVie+MCAN~\cite{nguyen2020revisiting} (A similar method was introduced by~\citet{jiang2020defense}), which uses a modulated convolutional bottleneck for the image backbone.
Other methods such as Pythia~\cite{jiang2018pythia}, VLBERT~\cite{su2019vl} and MMBT~\cite{kiela2019supervised} can benefit from our method, as well as more recent work such as UNITER~\cite{chen2020uniter}, which use the alignment task for pre-training their models.
Oscar~\cite{li2020oscar}, while using extensive data for pre-training, also introduce the use of objects' tags as additional inputs.
Because the architecture of UNITER and Oscar is close to the ones we experiment with, we focus our work on three different models. We further note that MoVie+MCAN uses grid features instead of detection features, \ie no detector is needed (in oppose to most methods), which adds to our approach applicability.

\section{Method}
Our methodology is composed of two disjoint parts: 
(i) for a given external knowledge source~$\mathcal{K}$, consisting of $m$ modalities, we train a model (\ie, the \textit{Retriever)} to align between the different modalities. (ii) Given a knowledge source $\mathcal{K}$ and an alignment model, we train a downstream model~(\ie, the \textit{Reader)} by augmenting its inputs with extra data from from the knowledge source~$\mathcal{K}$.

\subsection{Cross-modal Alignment}
\label{sec:cross-modal}

Let $\mathcal{K}$ be consists of $m$ modalities, where each sample $s_i = (s^0_i, \dots,s^m_i)\in\mathcal{K}$ is a tuple of $m$ elements, corresponding to different modalities. 
Our alignment model encompasses $m$ encoders $E_m$, each composed of a feature-extraction module $F_m$, projection layer $P_m$, shared Transformer encoding layer $T$ with attention pooling, and a normalization layer $\mathcal{N}$:
\begin{align}
    E_m(x) = \mathcal{N}(T(P_m(F_m(x))))
\end{align}
From this point, we will consider the two-modality case of images and captions, as illustrated in Fig.~\ref{fig:cross-modal}.
For text and image feature extractors, $F_1$ and $F_2$, we use a pre-trained BERT model, and a pre-trained ResNet152 CNN backbone on ImageNet, respectively. The images are represented with convolutional grid features, chosen for robustness and speed, and these are flattened across the spatial dimension.
The projection layers $P_m$ project each modality to a constant dimension $d$. The projected sequences are then forwarded to a shared Transformer-encoding layer, and aggregated by an attention pooling layer, resulting in a vector representation for each modality. 
Finally, we normalize each vector using an $L2$ normalization, projecting all embeddings to the unit-sphere. Following~\citet{faghri2017vse++}, we only normalize the text embeddings due to image-caption imbalance (see Sec.~\ref{sec:datasets}).

We train our dense cross-modal retriever (DXR) using a contrastive loss, specifically using an in-batch hinge penalty with hard negatives~\cite{faghri2017vse++}. Given a batch, consisting of $b$ samples, $s_1\dots s_b$, for each sample $s_i$, let $s^1_i$ and $s^2_i$ be the positive pairs and $s^1_i$ and $s^2_{j\ne i}$ the negative pairs. We compute the pair-wise similarity between the two modalities, using a dot product:
\begin{align}
    {s^2_i}' =  \max_{j \ne i}\langle s^1_i, s^2_j\rangle,\quad
    {s^1_i}' =  \max_{j \ne i}\langle s^1_j, s^2_i\rangle
\end{align}
\begin{align}
    \mathcal{L}_{hard} = &\sum_i[\alpha + \langle s^1_i, {s^2_i}'\rangle -\langle s^1_i, s^2_i\rangle] \notag\\ &+\sum_i[\alpha + \langle {s^1_i}', s^2_i\rangle -\langle s^1_i, s^2_i\rangle]
\end{align}
where ${s^1_i}'$ and ${s^2_i}'$ are the hardest samples inside the batch, and  $\alpha$ is the margin constant.

\begin{figure*}[t]
    \centering
    \includegraphics[width=\linewidth]{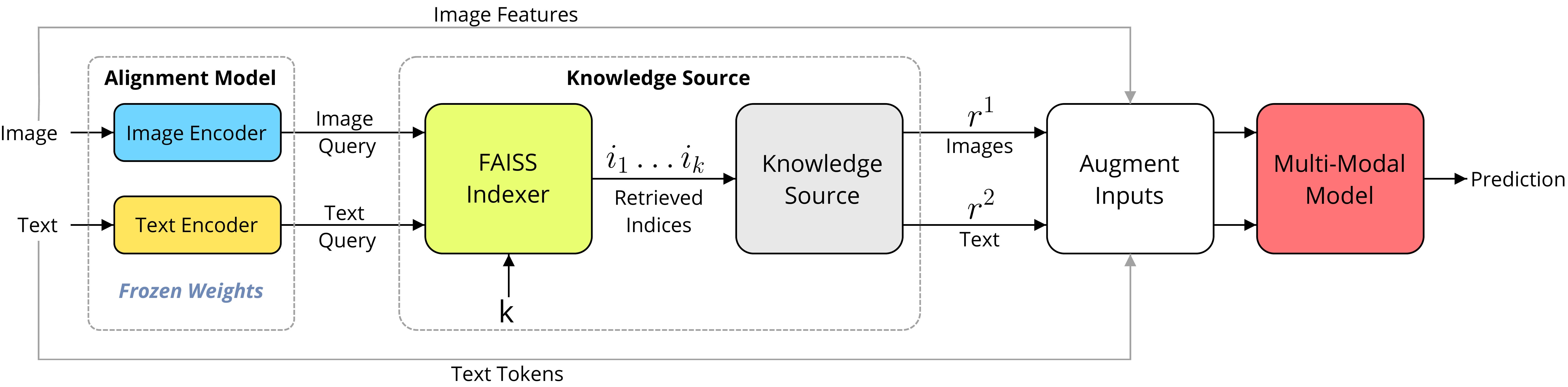}
    \caption{Illustration of our end-to-end framework. The trained cross-modal alignment is used to extract features as queries to a FAISS indexer. The $k$ retrieved indices are used to access data from the external knowledge source, and augment the input by appending each of the $k$ retrievals to the relative modality. For VQA, we only query the input image and retrieve $k$ captions.}
    \label{fig:e2e-modal}
\end{figure*}
\subsection{Indexing and Retrieving}
\label{sec:indexing_retrieving}
Following~\citet{lewis2020retrieval}, we use FAISS~\cite{JDH17} as our indexer platform for fast $\knn$ queries.

Given a knowledge source $\mathcal{K}$, we construct an index by computing the embeddings of each sample in $\mathcal{K}$ using some alignment model (the \textit{Retriever}), which can be trained on any arbitrary knowledge source.
We introduce with two variants: we either construct separate indices $I^m_\mathcal{K}$ for each of the modalities; or we construct one joint index $I_\mathcal{K}$ that encompasses all modalities and where a $\knn$ query will return a mixed modality result. 
Fig.~\ref{fig:e2e-modal} illustrates the two independent features of the alignment model and external knowledge source.

The retrieval process then consists of input query $q$, encoder $E_m$ and indexer $I_\mathcal{K}$ (or $I^m_\mathcal{K}$). $I_{\mathcal{K}}$ takes as an input an embedding query $e_q=E_m(q)$ and $k$, and returns the $k$-nearest indices $i_1 \dots i_k$, corresponding to the $k$-nearest embeddings. We then index data from $\mathcal{K}$, resulting in $m$ retrieval sets $r^m = (r^m_1 \dots r^m_{n_m})$, one for each modality, each consisting of varying number of samples $n_m$, where $\sum_{i=1}^m n_m = k$. When using $I^m_\mathcal{K}$, a single modality $m$ is returned, resulting in $r^m = (r^m_1 \dots r^m_k)$:
For simplicity, we define the retriever by $R(q, E_m, I_\mathcal{K}, k) := \{r^1, \dots, r^m\}$.

\subsection{End-to-End Fusion}
Let $M$ be any multi-modal reader model, applied to a specific downstream task that takes as an input $x=(x^1,\dots , x^m)$ consisting of $m$ modalities and outputs prediction $y$. The method augments the input $x$ by concatenating the retrieved samples to their corresponding input modalities, resulting in the augmented input $x'$:
\begin{align}
    x'=(x^1 \circ r^1_1 \circ \dots \circ r^1_{n_1},\dots , x^m \circ r^m_1 \circ \dots \circ r^m_{n_m})
\end{align}

The resulting end-to-end training of model $M$ is then defined by some loss function $\mathcal{L}$, minimizing $\mathcal{L}(M(x'),y)$, with the same hyperparameters as in the non-retrieval augmented case. Fig.~\ref{fig:e2e-modal} illustrates the complete model.

\subsection{Time-Complexity}
\label{sec:timecop}
As introduced in Sec.~\ref{sec:related}, we consider two types of retrievers, (i) methods such as ours, that use Maximum Inner Product Search (MIPS), where each modality is computed independently, and (ii) methods that have entangled computation of similarity between the different modalities, \eg they cannot compute an independent embedding.
Assuming a knowledge-source of size $N$, and a forward-pass with $O(1)$ time-complexity.
In type (i), the embeddings of the entire knowledge source need to be computed only once, as well as the embeddings of the query sample. In our experiments, we use FAISS with ``Hierarchical Navigable Small World'' search, and as shown by \citet{JDH17}, this searching method takes $O(AD[\log N]v)$, where $A$ and $v$ are constants, and $D$ is the degree of the graph. Therefore, the total time complexity of retrieving is $O(AD[\log N]v)$. 
On the other hand, methods of type (ii) must compute pairwise similarities between a query sample, and all samples in the dataset, resulting in a $O(N)$ run time for searching the most similar sample.

As a results, type (ii) methods are not applicable to our end-to-end fusion pipeline, where for each sample we query, an $O(N)$ is applied, which results in an inefficient and non-scalable method. On the other hand, our method does not impose significant overhead.

\section{Experiments}
In this section, we describe the two experimental settings of the alignment model and the end-to-end downstream task training and evaluation.  All models and experiments are implemented and performed with the MMF library~\citep{singh2020mmf}.

\subsection{Datasets}
\label{sec:datasets}
We use three common datasets for training and evaluating retrieval and VQA tasks.
Flickr-30K~\cite{young2014image} is composed of 30,000 images, with 5 captions each. Following~\citet{karpathy2015deep}, we use 1000 images for validation and 1000 images for testing. COCO~\cite{chen2015microsoft} is a well-known dataset that contains 120,000 images, with 5 captions each. We use the splits from~\citet{karpathy2015deep} as well, resulting in 80K images for training, 5K images for validation and 5K images for testing. Following~\citet{faghri2017vse++}, we add an additional 30K images for training, and uses the same 1K and 5K splits.
Conceptual Captions~\cite{sharma2018conceptual} is a dataset that contains image-caption pairs, composed of 3M samples for training and 100K for validation, which we use to test our retrieval model.

The proposed datasets differ in two major axes (i) Size - The largest knowledge-source we use is CC, which contains 3M image-caption pairs, while COCO and Flickr30K are smaller in one and two orders of magnitude, respectively. (ii) Domain gap - As shown in~\citet{singh2020we}, CC datasets differs in both visual and textual domain from the VQA task, while COCO has the best domain match in both. Flickr30K datasets, on the other hand, is very similar to COCO, but suffers from short number of samples in an order of magnitude compared to COCO.

\subsection{Cross-Modal Retrieval}
In the cross-modal retrieval task, we deal with two modalities: images and captions.
Bi-directional retrieval is evaluated, denoted as Text $\rightarrow$ Image and Image $\rightarrow$ Text, where the left-hand-side indicates the query and the other indicates the retrieved domain.
For fair comparison, we only report results for methods that use grid-features and vector representations, as noted in Sec~\ref{sec:indexing_retrieving} and~\ref{sec:timecop}. For a full comparison with other previous methods, please see Appendix~\ref{ap:retrieval}. Models are trained for 100K iterations with a warm-up of 2k iterations, batch size of 256, and Adam optimizer with learning-rate of $0.0001$ where the (pre-trained unimodal) feature encoder's learning-rate is multiplied by $0.1$. The hinge-loss margin hyperparameter $m$ is set to $0.2$.

\subsection{Downstream Tasks}
After training the alignment models for each dataset---Flickr30K, COCO and CC---we build indices for each, as defined in Sec~\ref{sec:indexing_retrieving}. Note that for COCO, we only use the training set for indexing, while for Flickr30K and CC, we use the entire set of train/val/test. This is done for fair comparison with the VQA task, which uses the COCO training-set images for training.
Our experiments focus on VQA as the downstream task, however we note that extension to other multi-modal tasks is straightforward.
The inputs of the VQA task are image and text tuples, and it is presented as a classification problem over a set of answers. In VQA, we observe that information regarding the content of the image, such as the amount, color and location of objects is very correlated with the question and answer. Therefore, captions serve as good auxiliary information, while similar/retrieved images (e.g., to which the question does not directly refer) are less informative. For that reason, we use the \textit{separate indices} variant, retrieving text captions from images to yield a cross-modal image to text translation. We experiment with all three datasets, evaluating different training and inference variants.

\section{Results}
\subsection{Cross-Modal Retrieval}
Tab.~\ref{tab:ret_coco} and~\ref{tab:ret_flickr} show retrieval results on COCO and Flickr30K, respectively, comparing similar methods that use grid-features and vector representations for the embedding space. Reported numbers correspond to Recall-at-1/5/10 on the \texttt{test-sets}. As can be seen, our method significantly outperforms previous work when trained on the same datasets. We also added the results for CLIP and ALIGN. We note that both of these methods use significantly larger amounts of external training data (0.4 and 1.8 Billion resp.). We refer to Appendix~\ref{ap:retrieval} for a comparison with additional methods.

While CC is not commonly used in the retrieval literature, we use it for our downstream task. Using DXR, we obtain the following results for CC: R@1: $25.1$ R@5: $50.1$ and R@10: $61.9$ for Text $\rightarrow$ Image, and R@1: $25.4$ R@5: $50.9$ and R@10: $61.8$ for Image $\rightarrow$ Text. The alignment model trained on CC is used for training in the downstream VQA task.
We notice that performance degrades as the dataset size increases, which could affect the downstream task since we query from the entire dataset.

\begin{table*}[t]
    \small
    \centering
    \begin{tabular*}{\linewidth}{@{\extracolsep{\fill}}lc@{~~}c@{~~}cc@{~~}c@{~~}c|c@{~~}c@{~~}cc@{~~}c@{~~}c}
        \toprule
         & \multicolumn{6}{c}{COCO 1K} & \multicolumn{6}{c}{COCO 5K}\\
         & \multicolumn{3}{c}{Text $\rightarrow$ Image} & \multicolumn{3}{c}{Image $\rightarrow$ Text} & \multicolumn{3}{c}{Text $\rightarrow$ Image} & \multicolumn{3}{c}{Image $\rightarrow$ Text}\\
        Method & R@1 & R@5 & R@10 & R@1 & R@5 & R@10 & R@1 & R@5 & R@10 & R@1 & R@5 & R@10\\
        \midrule
        DPC & 47.1 & 79.9 & 90.0 & 65.6 & 89.8 & 95.5 & 25.3 & 53.4 & 66.4 & 41.2 & 70.5 & 81.1\\
        VSE++ & 52.0 & 83.1 & 92.0 & 64.6 & 89.1 & 95.7 & 30.3 & 59.1 & 72.4 & 41.3 & 69.2 & 81.2\\
        CMPM & 44.6 & 78.8 & 89.0 & 56.1 & 86.3 & 92.9 & 22.9 & 50.2 & 63.8 & 31.1 & 60.7 & 73.9\\
        \textbf{DXR} & \textbf{56.8} & \textbf{88.2} & \textbf{94.9} & \textbf{67.0} & \textbf{93.0} & \textbf{97.6} & \textbf{33.9} & \textbf{64.9} & \textbf{77.4} & \textbf{44.9} & \textbf{75.2} & \textbf{84.7}\\
        \midrule
        CLIP$^\dagger$ & - & - & - & - & - & - & 37.8 & 62.4 & 72.2 & 58.4 & 81.5 & 88.1\\
        ALIGN$^\dagger$ & - & - & - & - & - & - & \underline{45.6} & \underline{69.8} & \underline{78.6} & \underline{58.6} & \underline{83.0} & \underline{89.7}\\
        \bottomrule
    \end{tabular*}
    \caption{Retrieval results for COCO, comparing only methods that use raw images as input, and vector representations for the embedding space. We denote by $\dagger$ methods that train on substantial amount of novel data. Additional methods can be found in Appendix~\ref{ap:retrieval}.}
    \label{tab:ret_coco}
\end{table*}
\begin{table}[t]
    \small
    \centering
    \begin{tabular*}{\linewidth}{@{\extracolsep{\fill}}lc@{~~}c@{~~}c@{~~}c@{~~}c@{~~}c}
        \toprule
         & \multicolumn{3}{c}{Text $\rightarrow$ Image} & \multicolumn{3}{c}{Image $\rightarrow$ Text}\\
        Method & R@1 & R@5 & R@10 & R@1 & R@5 & R@10\\
        \midrule
        RRF & 35.4 & 68.3 & 79.9 & 47.6 & 77.4 & 87.1\\ 
        CMPM & 37.3 & 65.7 & 75.5 & 49.6 & 76.8 & 86.1\\
        DPC & 39.1 & 69.2 & 69.2 & 55.6 & 81.9 & 89.5\\
        VSE++ & 39.6 & 69.6 & 79.5 & 52.9 & 79.1 & 87.2\\
        \textbf{DXR} & \textbf{50.6} & \textbf{78.8} & \textbf{86.7} & \textbf{65.1} & \textbf{87.3} & \textbf{92.6}\\
        \midrule
        CLIP$^\dagger$ & 68.7 & 90.6 & 95.2 & 88.0 & \underline{98.7} & 99.4 \\
        ALIGN$^\dagger$ & \underline{75.7} & \underline{93.8} & \underline{96.8} & \underline{88.6} & \underline{98.7} & \underline{99.7} \\
        \bottomrule
    \end{tabular*}
    \caption{Retrieval results for Flickr30K, comparing only methods that use raw images as input, and vector representations for the embedding space. We denote by $\dagger$ methods that train on substantial amount of novel data. Additional methods can be found in Appendix~\ref{ap:retrieval}.}
    \label{tab:ret_flickr}
\end{table}
\begin{table}[t]
    \small
    \centering
    \begin{tabular*}{\linewidth}{@{\extracolsep{\fill}}ll@{~~~}cc}
    \toprule
    Knowledge & \multirow{2}{*}{Training Type} & \multirow{2}{*}{Visual BERT} & \multirow{2}{*}{ViLBERT}\\
     Source &  &  & \\
    \midrule
    Flickr30K & XTRA 10-C & 66.77 & 67.32\\
    \midrule
    \multirow{3}{*}{CC} & PT & 64.34 & 68.14\\
    & XTRA-10C & 67.49 & 67.37\\
    & PT + XTRA-10C & 67.53 & 69.17\\
    \midrule
    \multirow{3}{*}{COCO} & PT & 64.54 & 67.58\\
    & XTRA-10C & \textbf{68.98} & 69.07\\
    & PT + XTRA-10C & 67.71 & \textbf{69.90}\\
    \midrule
    \multicolumn{2}{c}{Vanilla} & 63.54 & 67.56\\
    \midrule
    \multicolumn{2}{c}{5-GT} & \bgray{69.61} & \bgray{71.50}\\
    \bottomrule
    \end{tabular*}
    \caption{VQA Results for Visual-BERT and ViLBERT models on COCO \texttt{val-set}. \textbf{Vanilla} - models use pre-trained BERT model. \textbf{PT} - Pre-Training with the knowledge source. \textbf{XTRA-10C} - training via our method using the knowledge source indicated and alignment model trained on that knowledge source, using 10 retrieved captions.}
    \label{tab:main_vqa}
\end{table}
\begin{table}[t]
    \small 
    \centering
    \begin{tabular*}{\linewidth}{@{\extracolsep{\fill}}c@{~~~}cc@{~~~}c@{~~~}c|c@{~~~}c}
        \toprule
        \multirow{3}{*}{Flickr30K} & \multirow{3}{*}{CC} & \multicolumn{3}{c|}{COCO} & \multirow{3}{*}{Vanilla} & \multirow{3}{*}{5-GT}\\
        &&\multirow{2}{*}{\texttt{val}}&\multicolumn{2}{c|}{\texttt{test}}&&\\
        &&&\texttt{dev}& \texttt{std} &&\\
        \midrule
        69.70 & 69.02 & \textbf{71.52} & \textbf{72.80} & \textbf{73.12} & 71.16 & \bgray{71.80}\\
        \bottomrule
    \end{tabular*}
    \caption{VQA Results for MoVie+MCAN model, using XTRA-10C training type.}
    \label{tab:movie_mcan}
\end{table}

\subsection{Visual Question Answering}
Our main results show performance on the VQA \texttt{val-set}, experimenting with three common VQA methods: VisualBERT~\cite{li2019visualbert}, ViLBERT~\cite{lu2019vilbert}, and the currently winner of the VQA 2.0 challenge, Movie+MCAN~\citep{nguyen2020revisiting}, each along with three different knowledge sources (COCO, CC and Flickr30K). Following \citet{jiang2020defense}, we use the \texttt{val-set} to assist in our exhaustive ablation studies, however we report our final result on the VQA \texttt{test-dev} and \texttt{test-std} splits.

\begin{figure*}[t]
    \centering
    \begin{tabular}{@{}cc}
        \includegraphics[width=0.48\linewidth]{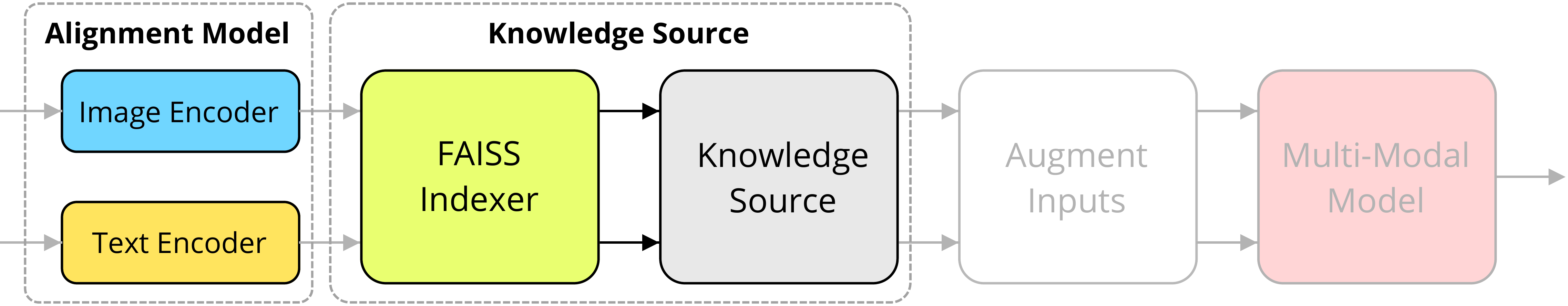} &
        \includegraphics[width=0.48\linewidth]{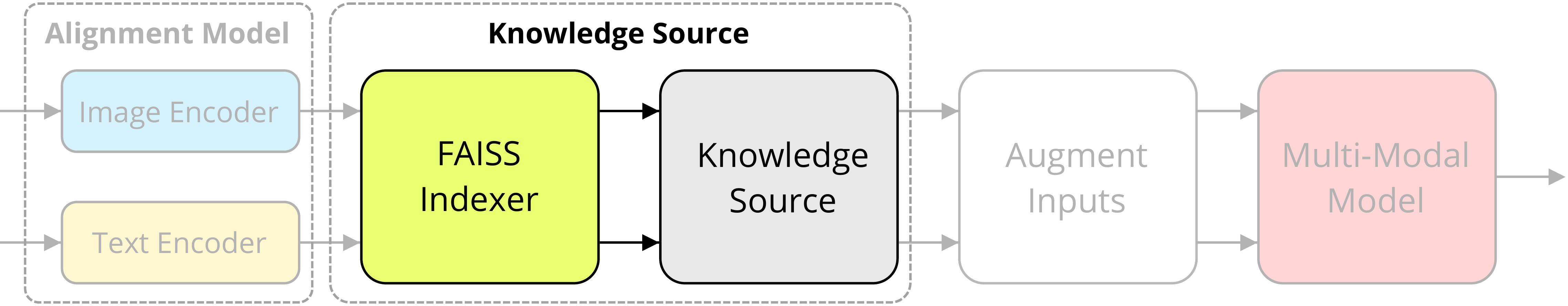} \\
        (a) & (b)
    \end{tabular}
    \caption{Two Hot-Swap configurations of the knowledge source during inference. \textbf{(a)} both the alignment model and the knowledge source are replaced with new ones built using a new dataset. 
    \textbf{(b)} only the knowledge source is replaced, and the indexer is built using the old alignment model.}
    \label{fig:hs}
\end{figure*}
\begin{figure*}[t]
    \centering
    \begin{tabular}{@{}cccc}
    \includegraphics[width=0.23\linewidth]{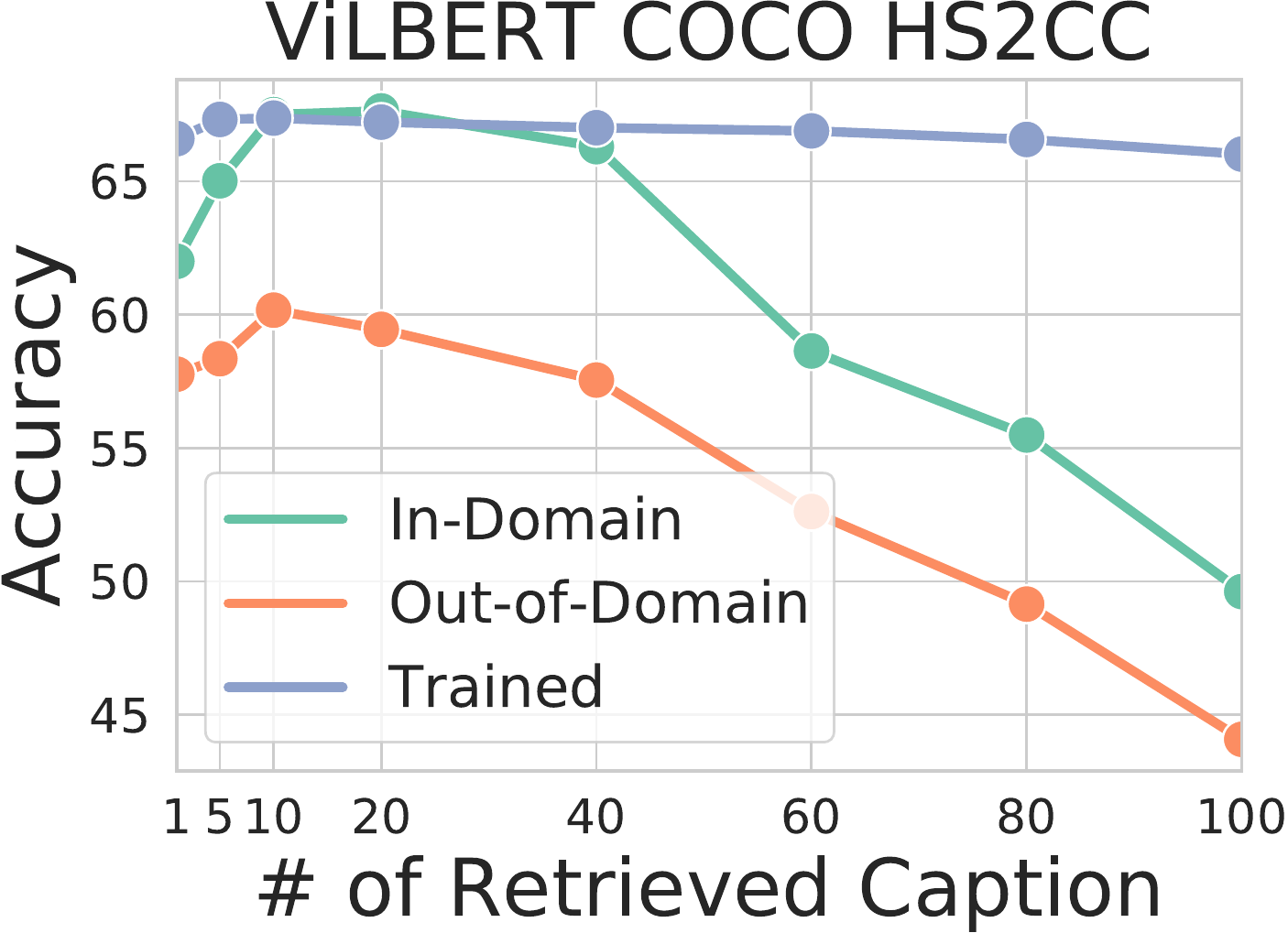} &
    \includegraphics[width=0.23\linewidth]{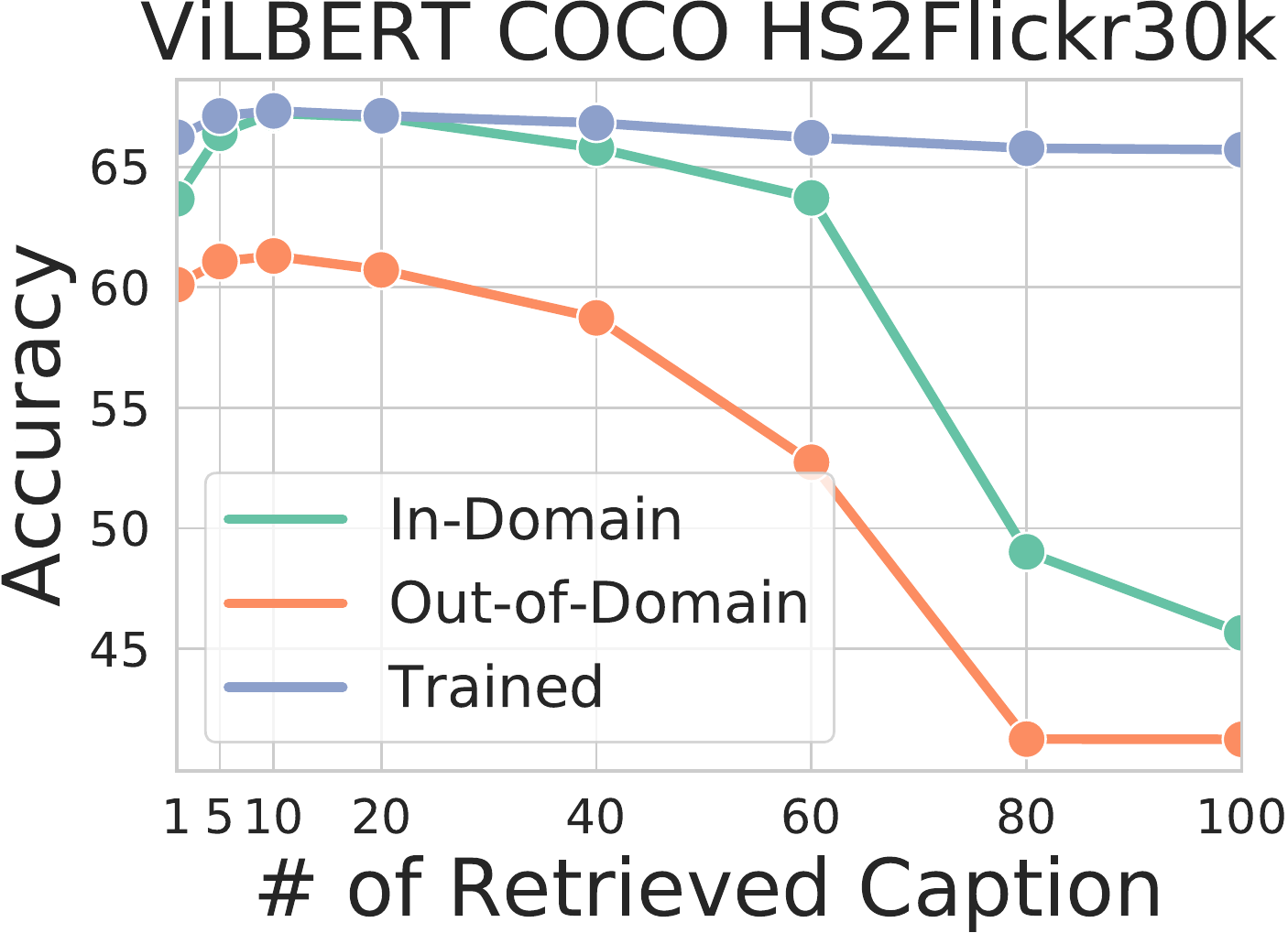} &
    \includegraphics[width=0.23\linewidth]{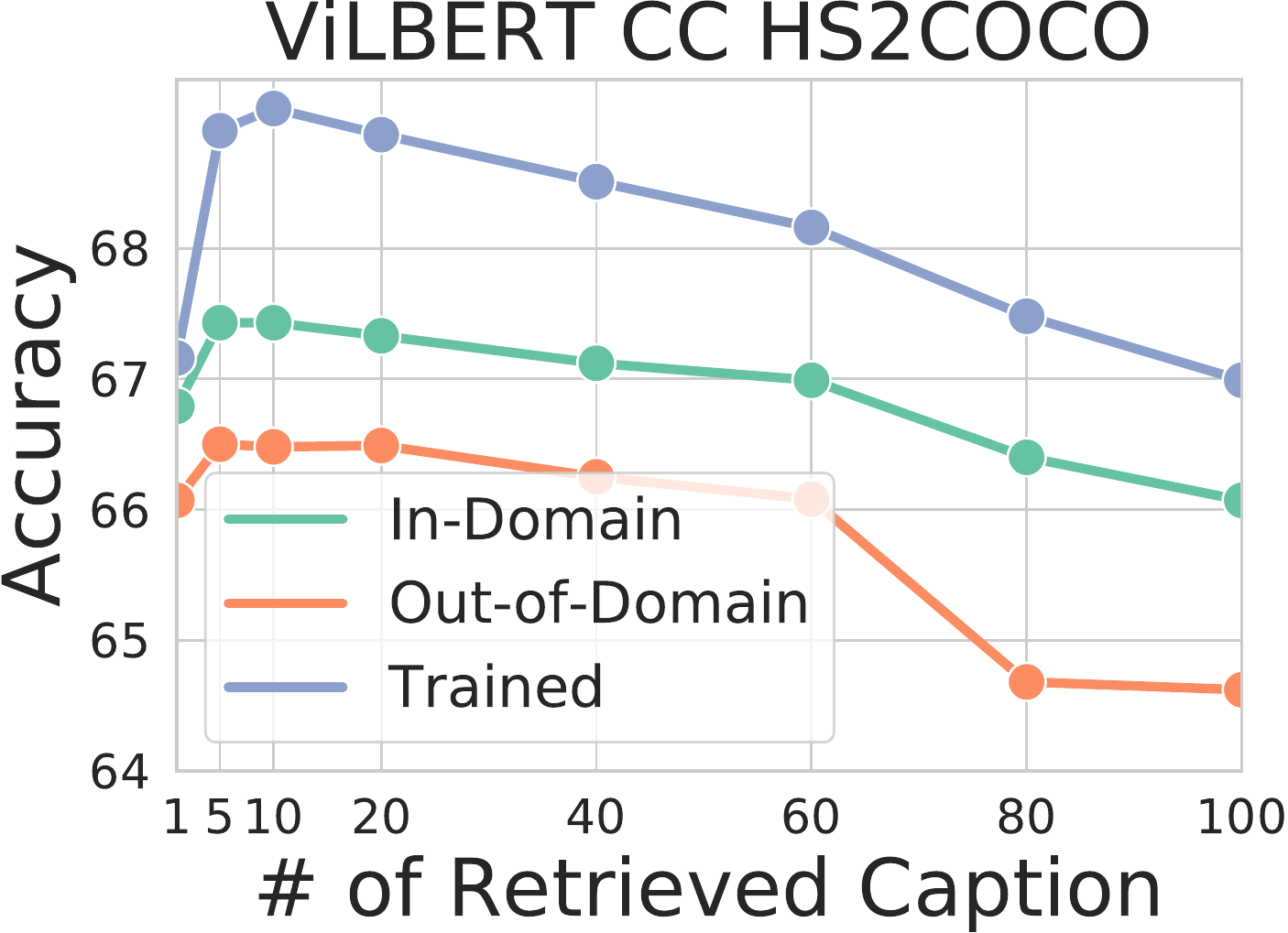} &
    \includegraphics[width=0.23\linewidth]{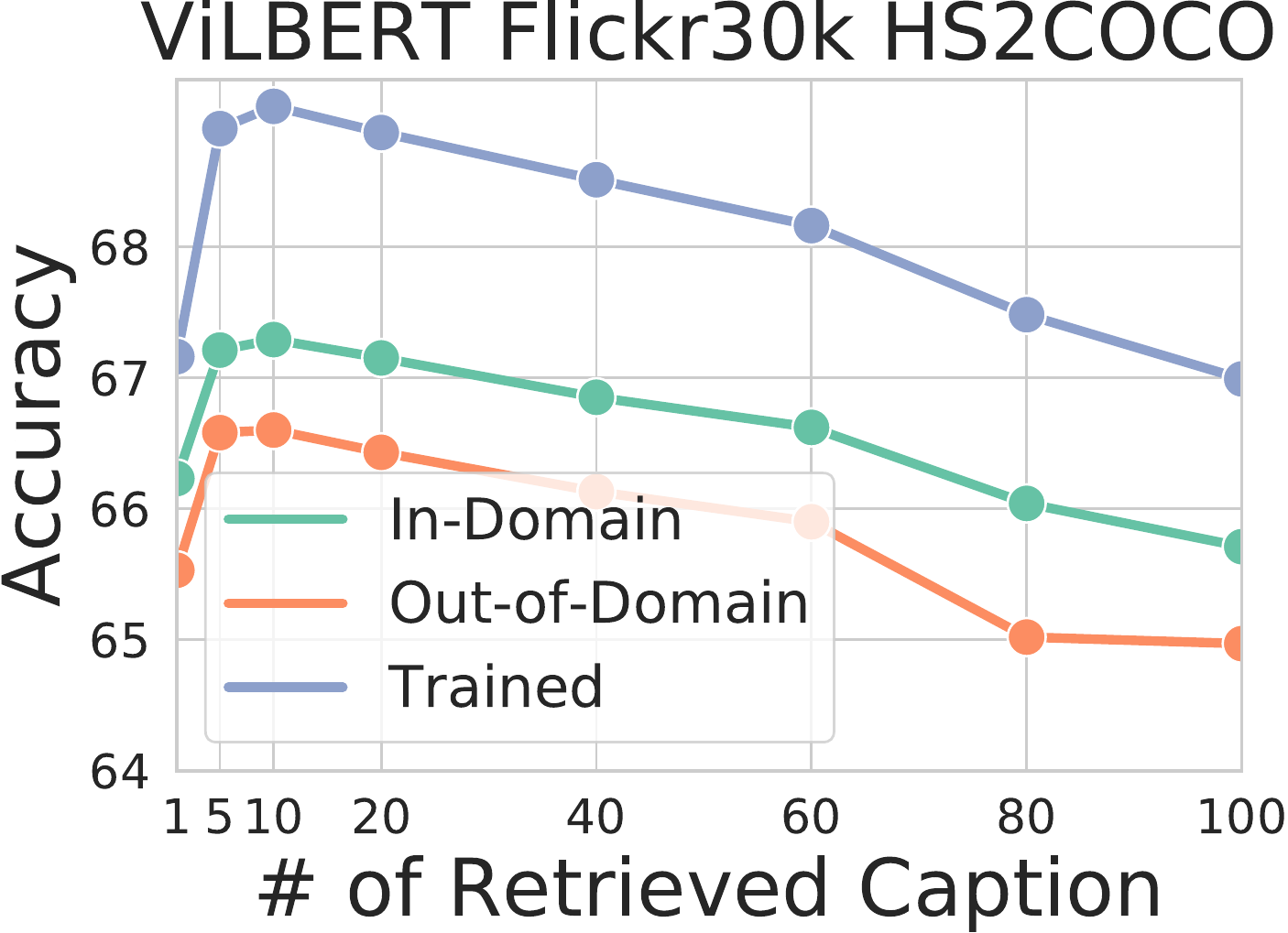} \\
    \includegraphics[width=0.23\linewidth]{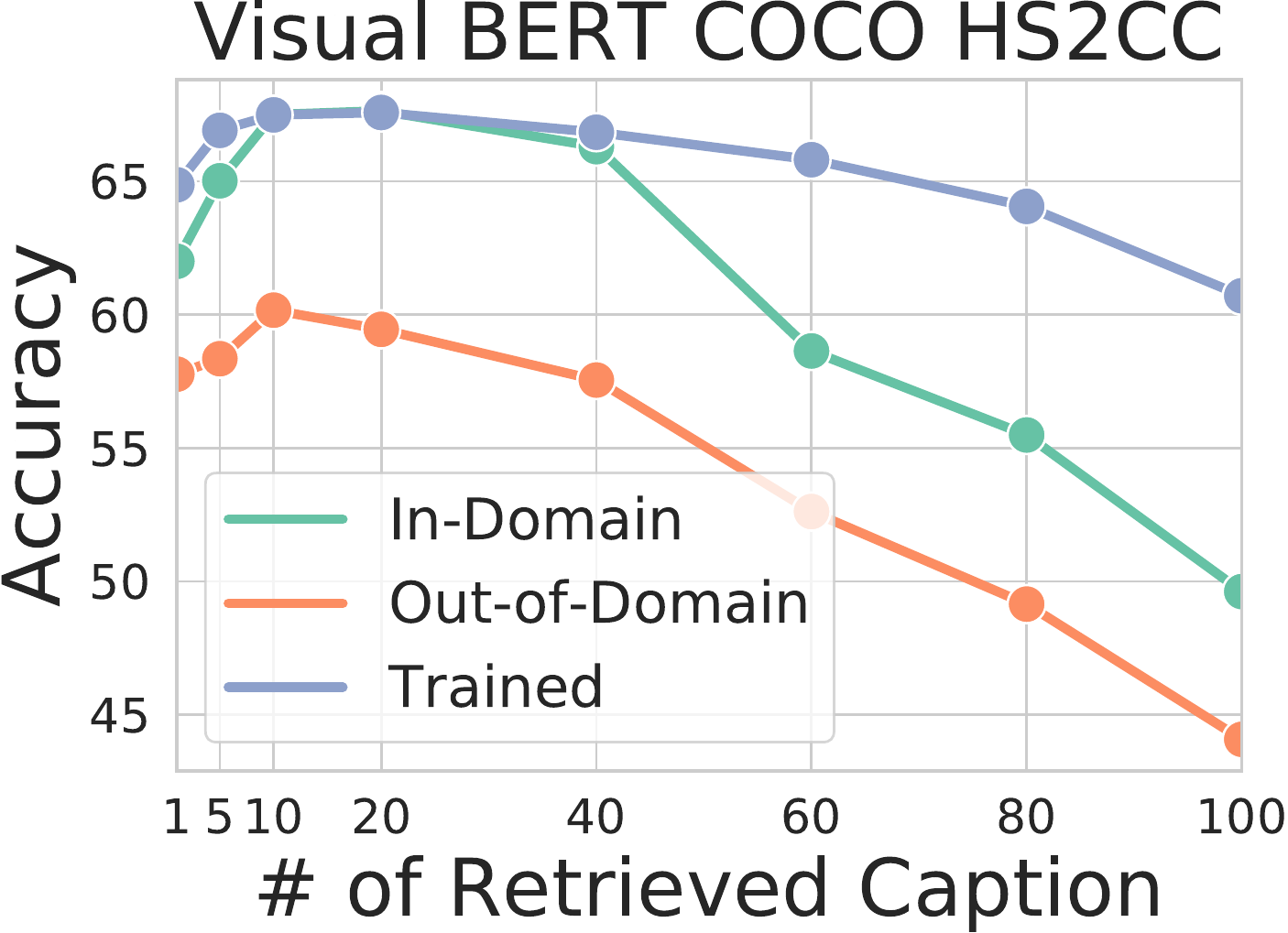} &
    \includegraphics[width=0.23\linewidth]{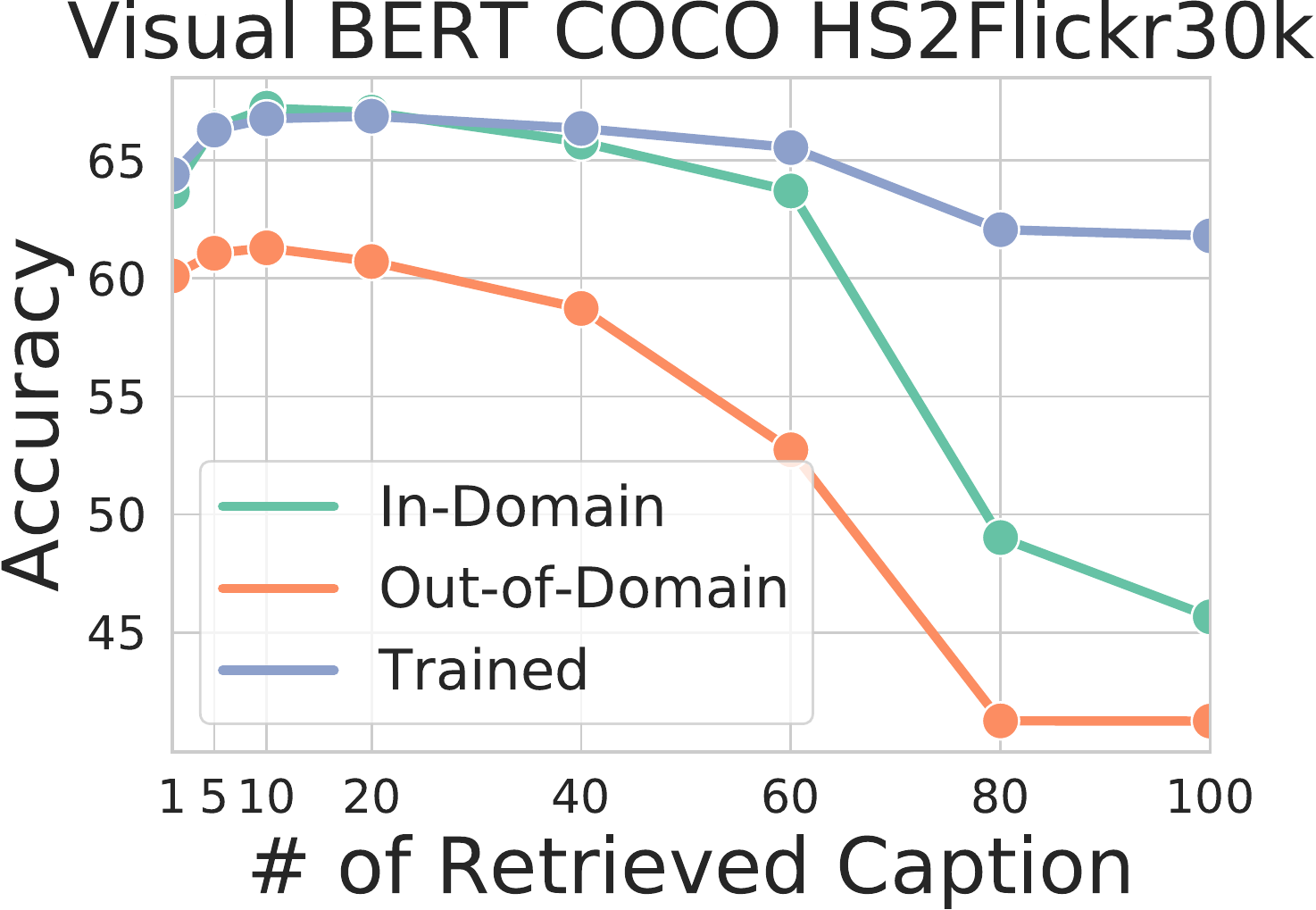} &
    \includegraphics[width=0.23\linewidth]{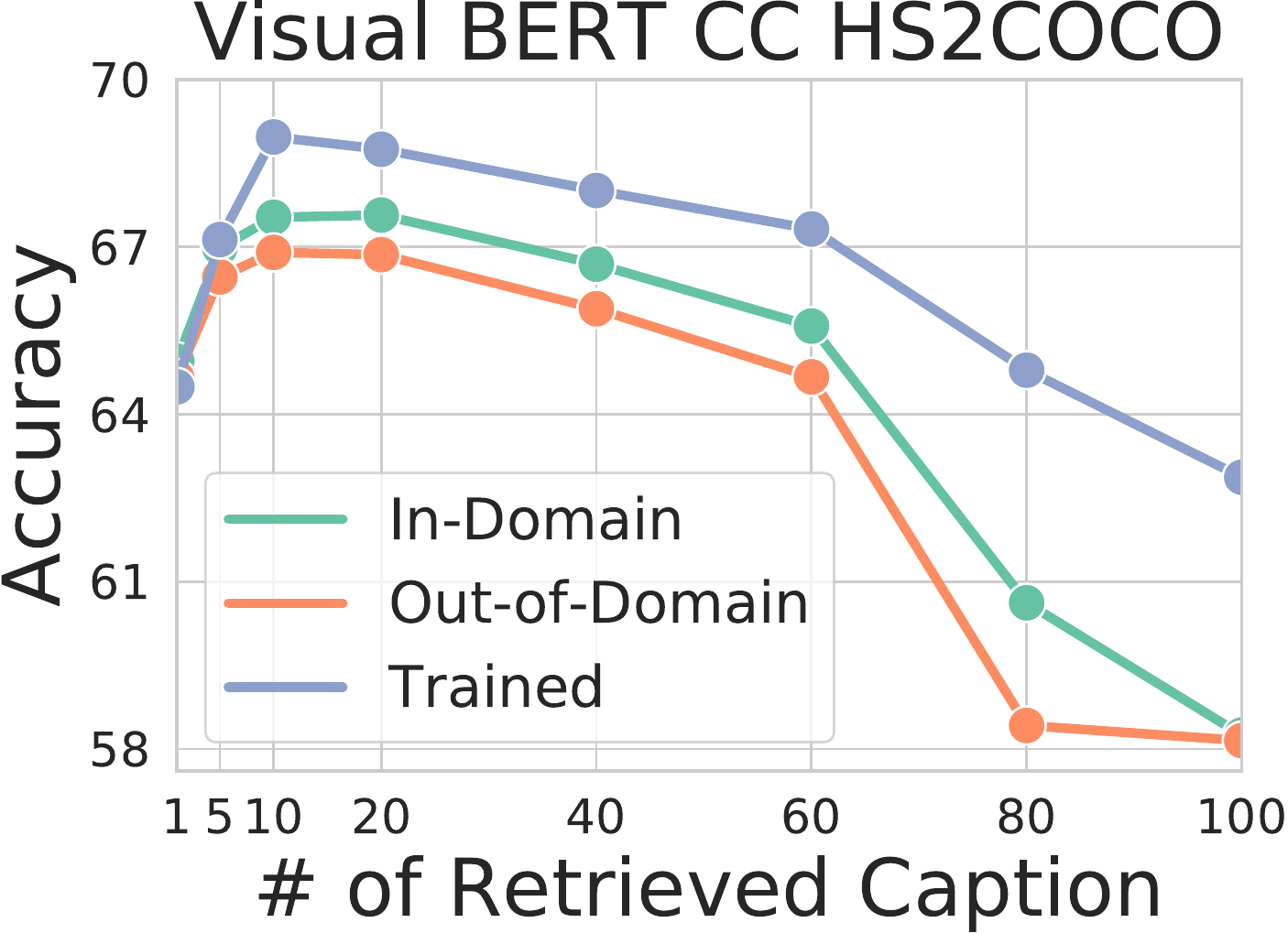} &
    \includegraphics[width=0.23\linewidth]{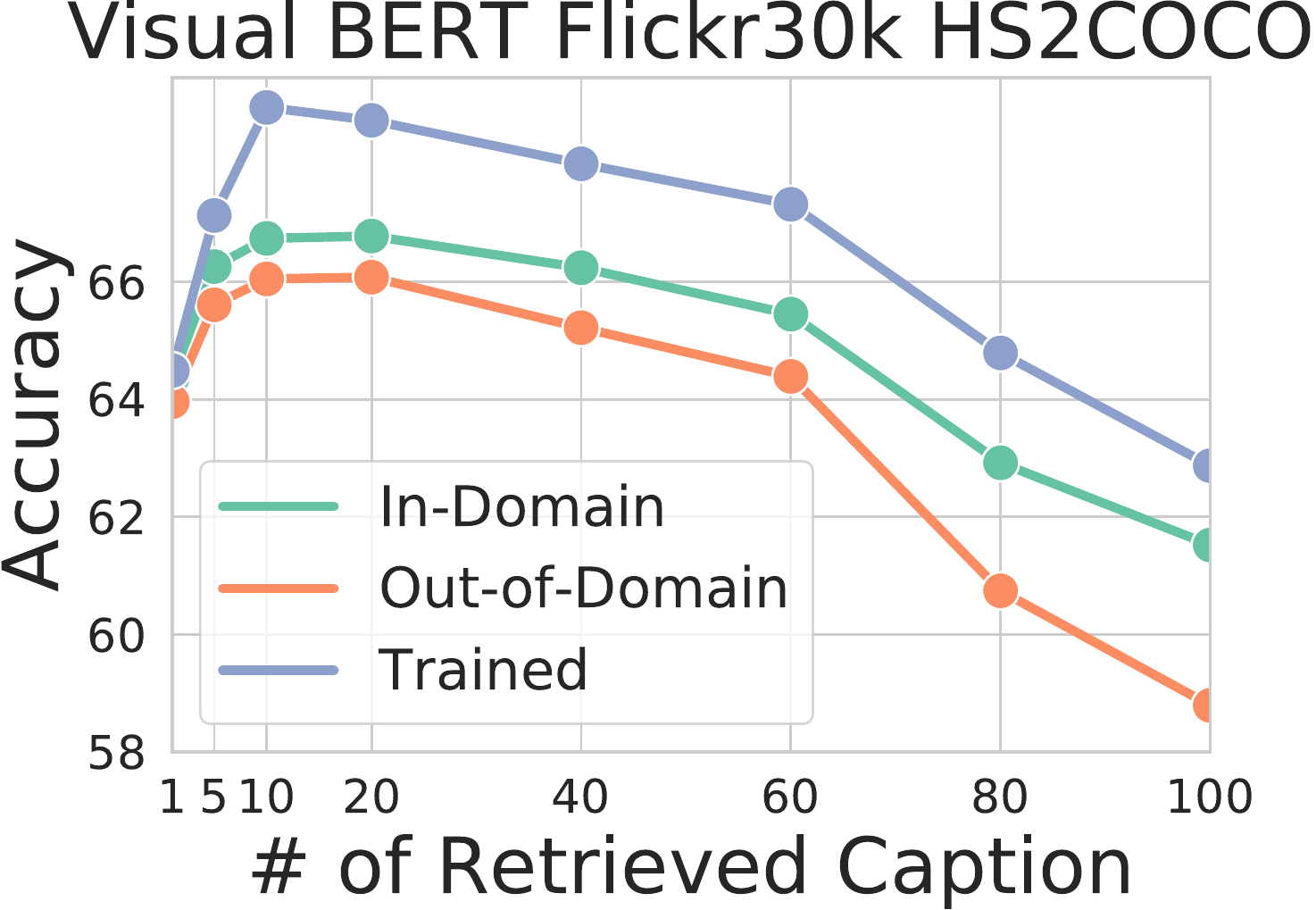}
    \end{tabular}
    \caption{Hot-Swap results. Each row corresponds to a different reader model. Each graph shows \textbf{(a)} Training with different amount of retrieved captions. \textbf{(b)} Using the trained model with 10-cap, we inference with different amount of captions. \textbf{(c)} Hot swapping between knowledge sources.}
    \label{fig:applications}
\end{figure*}
Tab.~\ref{tab:main_vqa} and~\ref{tab:movie_mcan} summarize four different training settings: (i) \textbf{vanilla} - models using pre-trained BERT; (ii) \textbf{PT} - Task agnostic pre-training with the knowledge source dataset (using masked language modeling); (iii) \textbf{5-GT} - training with the 5 ground truth captions from COCO; (iv) \textbf{XTRA-10C} - training via our method, using the knowledge source indicated and alignment model trained on that source, using 10 retrieved captions.
We see that using the five ground truth (GT) COCO captions as additional data (bottom row of Tab.~\ref{tab:main_vqa}), sets a soft upper bound for our approach. On one hand, the GT captions contain relevant information about the content of the image; on the other hand, other captions from the knowledge source may additionally serve as rich, useful descriptions. 
We also see that our method increases performance across all baselines, even with respect to pre-training. This suggests that our method serves as a good alternative for pre-training.

For the MoVie+MCAN model, we also report results for \texttt{test-dev} and \texttt{test-std} for COCO as our KS, setting our best model to be Movie+MCAN+XTRA-10C, obtaining a score of \textbf{73.12} for \texttt{test-std} (with single model performance). \citet{jiang2020defense} reported $72.71$ on \texttt{test-dev} while training on the same data as our method (COCO \textit{train+val}), while our approach achieves \textbf{72.8}. 
\citet{nguyen2020revisiting} on the other hand, train with a larger VQA dataset using COCO and Visual Genome~(VG)~\cite{krishna2017visual}, reporting $72.91$ on \texttt{test-dev}.

\label{sec:applications}
\begin{figure*}[t]
    \renewcommand{\arraystretch}{1} 
    \centering
    \begin{tabular*}{\linewidth}{@{}c@{}|L@{~}L@{~}L}
    \toprule
    Query& \multicolumn{3}{c}{Retrieved Captions}\\
    Image & No Hotswap  & Flickr30K Hotswap & CC Hotswap\\
    \rowfont{\fontsize{8}{5} \selectfont}
    COCO \texttt{val-set} & COCO \texttt{train-set} & \texttt{train+val+test sets}& \texttt{train+val sets}\\
    \midrule
    \raisebox{3px}{\multirow{5}{*}{\includegraphics[height=38px]{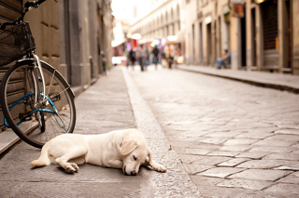}}}
    \rowfont{\fontsize{7}{10} \selectfont}
    &A dog that is lying down on a sidewalk  & A dog asleep on the streets  & A dog lies down on a cobblestone street\\
    &A dog with a muzzle on is lying on the sidewalk  & A tan male bulldog sleeping on a sidewalk & The dog is lying on the cobblestone street \\
    &A happy stray puppy lies in the street  & Cute dog sleeping on the sidewalk  & A dog laying on the side of the street\\
    &A dog is laying and resting on a walkway  & A dog lying on the sidewalk & A dog with a collar on lying on the street\\
    \bottomrule
    \end{tabular*}
    \caption{Sample top-4 result for ``in-domain'' Hot-Swap. The model was trained using COCO as the knowledge source, and 10 retrieved captions. Left - Query image from VQA \texttt{val-set}.
    Columns refer to the different hot-swaps, showing retrieved captions.}
    \label{fig:inf_ret}
\end{figure*}
\begin{figure*}[t]
    \centering
    \begin{tabular}{@{}cccc}
    \includegraphics[width=0.23\linewidth]{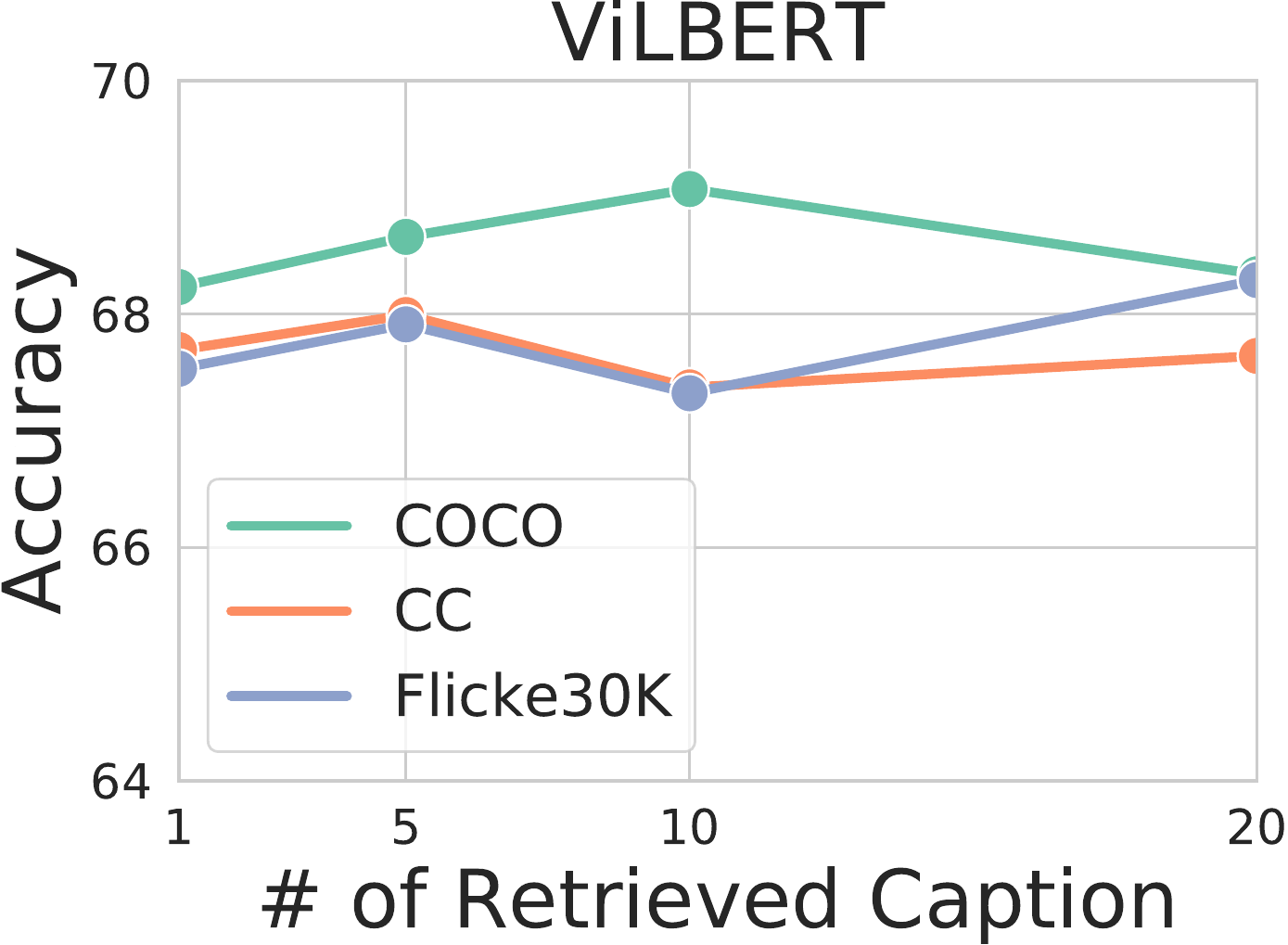} &
    \includegraphics[width=0.23\linewidth]{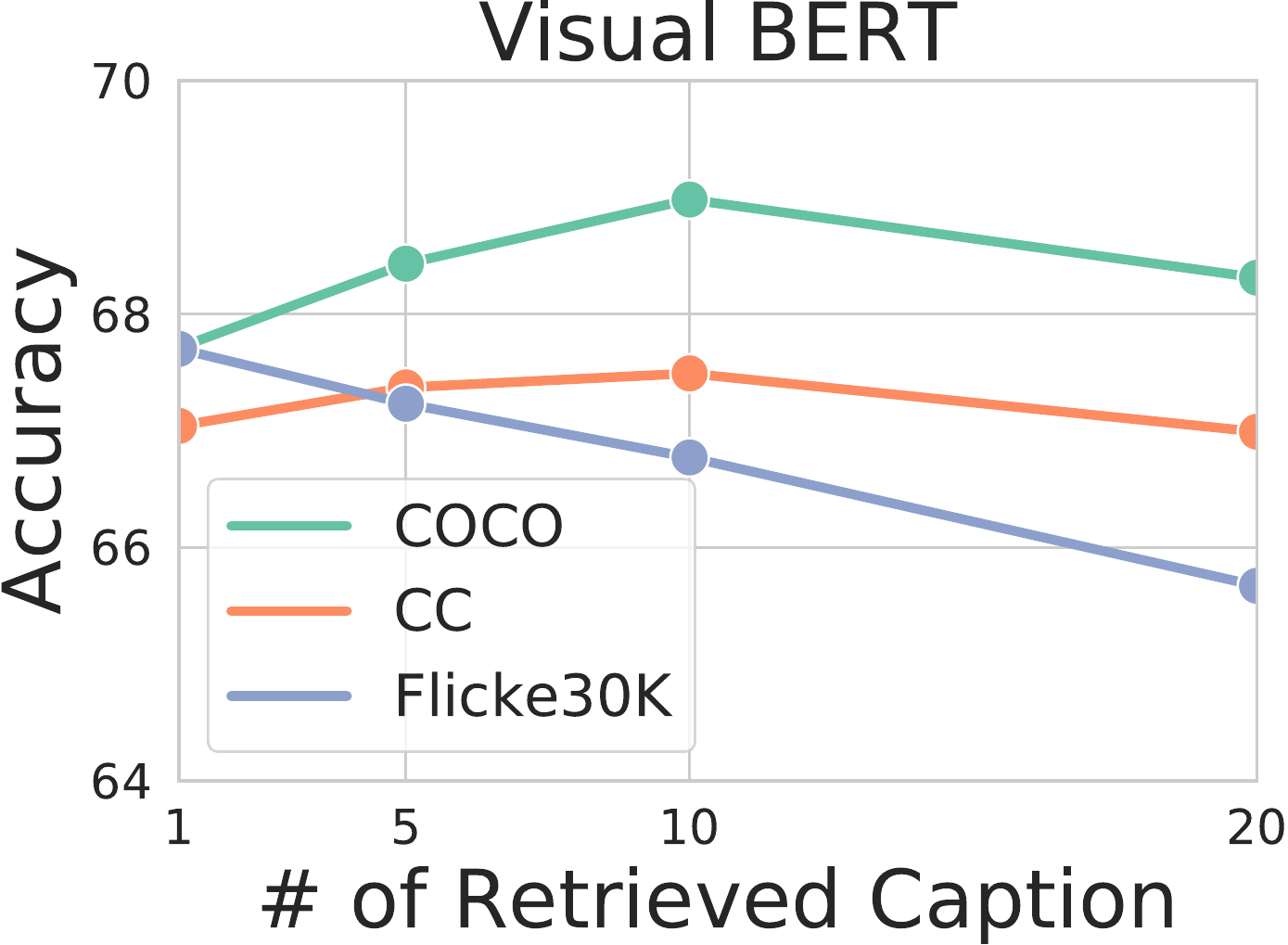} &
    \includegraphics[width=0.23\linewidth]{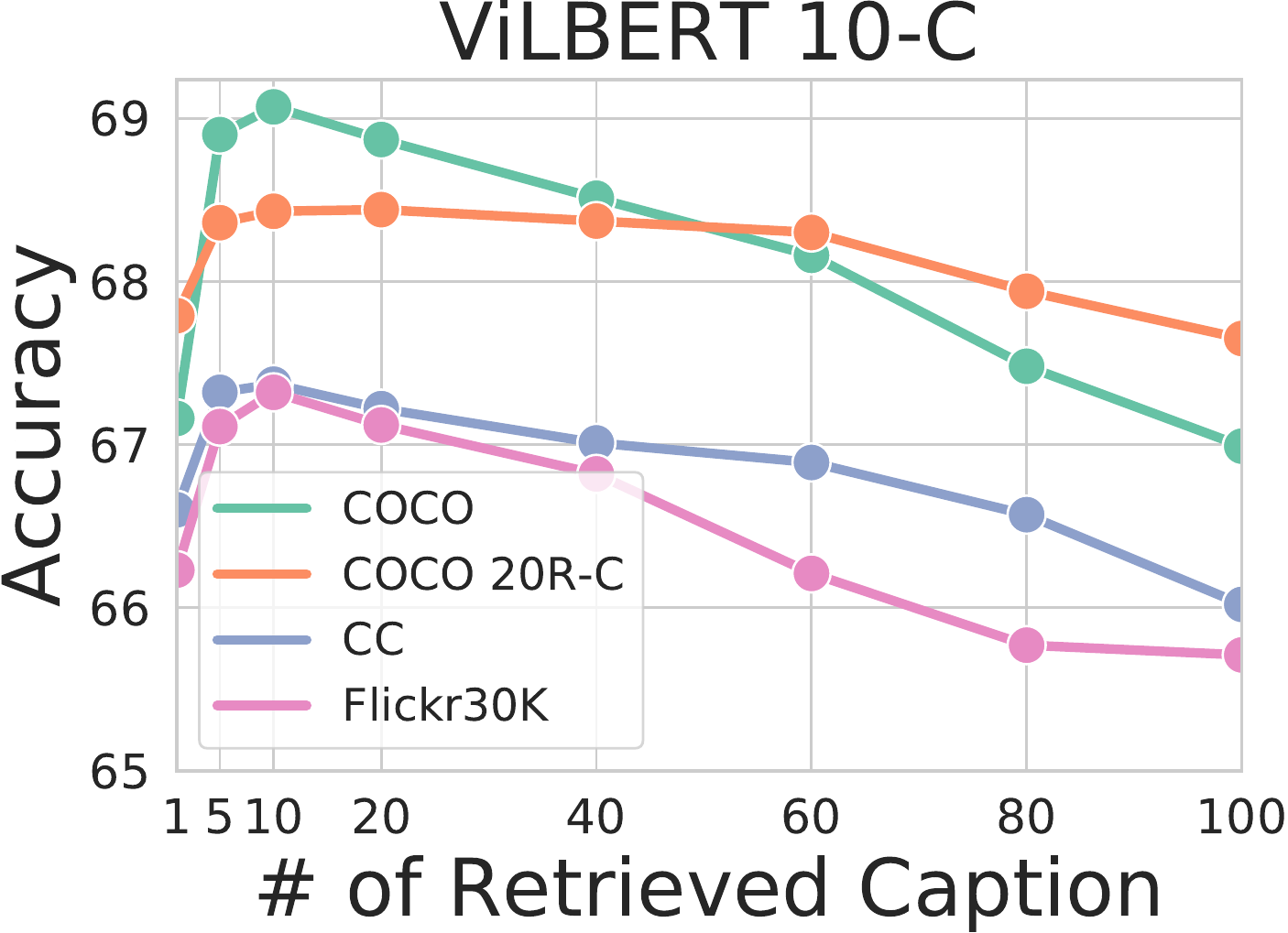} &
    \includegraphics[width=0.23\linewidth]{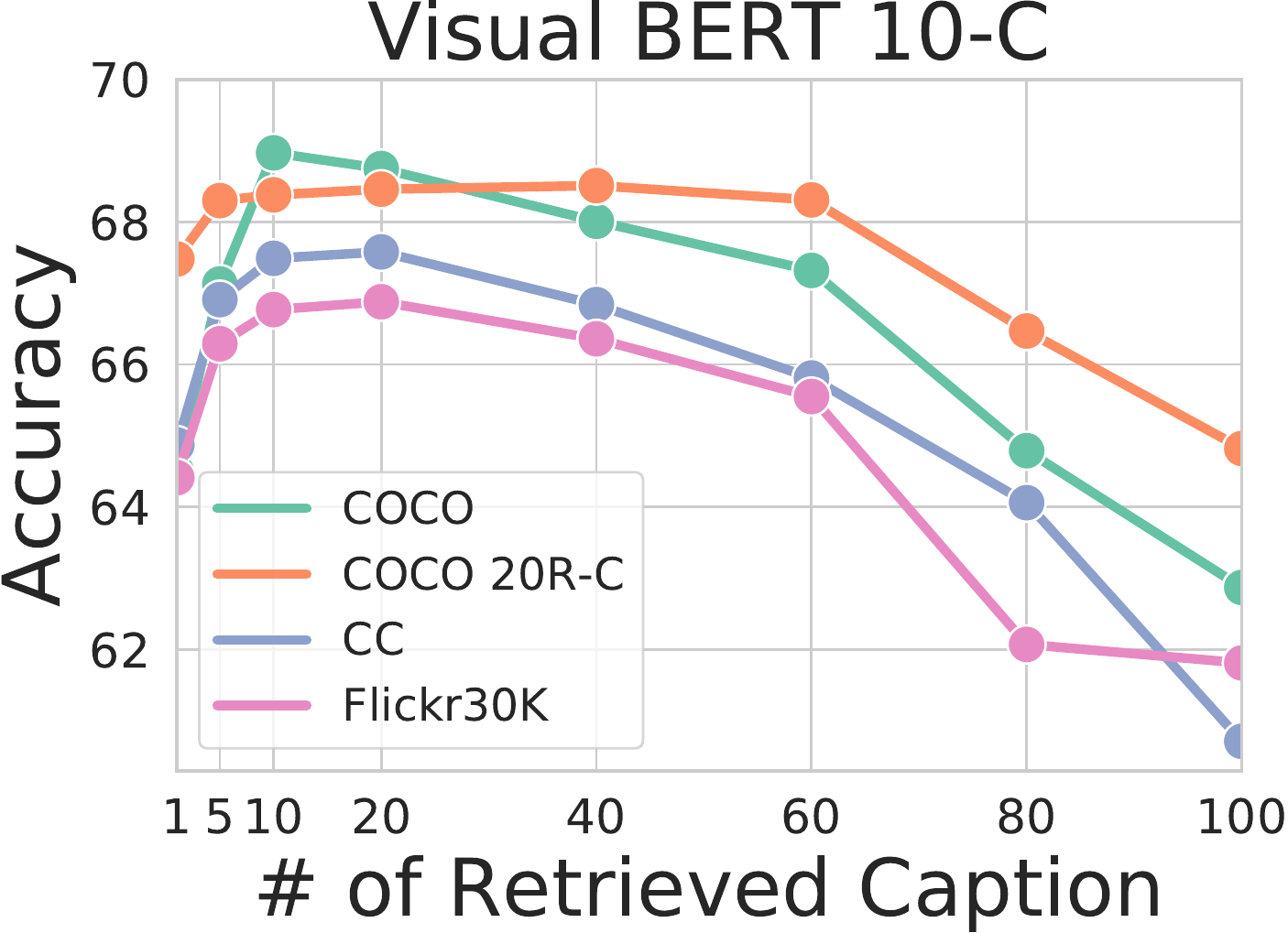} \\
    \multicolumn{2}{c}{(a)} & \multicolumn{2}{c}{(b)}
    \end{tabular}
    \caption{Ablation study of our method. \textbf{(a)} - Training with different amount of retrieved captions. \textbf{(b)} - Using the trained model with 10-cap, we inference with different amount of captions.}
    \label{fig:ablation}
\end{figure*}
\begin{table}[t]
    \small
    \centering
    \begin{tabular*}{\linewidth}{@{\extracolsep{\fill}}l@{~~~~}cc}
    \toprule
    Knowledge & \multirow{2}{*}{Visual BERT} & \multirow{2}{*}{ViLBERT}\\
    Source & & \\
    \midrule
    COCO &  58.77 (68.98) \down{10.21} & 45.60 (69.07) \down{23.47}\\
    CC & 63.15 (67.49) \down{4.34} &  63.50 (67.37) \down{3.87}\\
    Flickr30K & 61.86 (66.77) \down{4.91} & 59.34 (67.32) \down{7.98}\\
    \bottomrule
    \end{tabular*}
    \caption{VQA performance using models trained with 10 retrieved caption, and evaluating without any retrievals ("unplugged"). The highest decrease in performance occurs for the in-domain (COCO) knowledge source where retrieved examples are most informative.}
    \label{tab:unplugged}
\end{table}

\subsection{Hot Swap}
Our method is devised such that querying and retrieving from the knowledge source is independent of the downstream model, enabling the swap of the alignment model and/or knowledge source during inference. This affords interesting explorations.  We describe two forms of ``hot swapping'': (i) the entire knowledge source with its trained alignment model are replaced with a new one and corresponding alignment model -- we refer to this as ``out-of-domain''; (ii) the knowledge source used for retrieving is swapped, but the alignment model remains the same as was originally trained with the downstream model. In this case, we build a new retriever for the new knowledge source, using the original alignment model -- we call this ``in-domain''. ``in/out-of-domain'' refers to the alignment domain with which the downstream model was trained. 
Fig.~\ref{fig:hs} illustrates the two cases.

In Fig.~\ref{fig:applications} we show different inference results for hot swapping. All models in this experiment are trained using 10 retrieved cations. The title of each graph represents the trained model, followed by the trained knowledge source and the knowledge source to which we swap. In addition, we show inference results for training with the swapped knowledge source, \eg~training with CC knowledge source and alignment model from scratch, using 10 retrievals. As can be seen, ``in-domain'' hot swapping performance is significantly higher than ``out-of-domain''. We hypothesize that the reader model has learned an implicit structure of the alignment space.
Surprisingly, when training with COCO as the knowledge source, ``in-domain'' hot swapping performs similarly, for the same amount of trained retrievals (10), as training with an alternative knowledge source and alignment model. On the other hand, we observe that this suffers from a decrease in generalization due to different amounts of retrieval during inference-time. In the other direction, hot swapping to COCO from CC or Flickr30K does not result in the same performance as training with COCO as the knowledge source and alignment model, yet, performance and generalization do not degrade.
Qualitative results of ``in-domain'' hot swapping are presented in Fig~\ref{fig:inf_ret}. As can be observed, novel information such as the fact that the image shows a ``cobblestone street'' is retrieved from CC without having to train the alignment model on that source.

\subsection{Ablation Study}
\label{sec:ablation}
In this study, we explore the use of different amounts of retrieval during training and inference, as well as doing inference without retrieving - which we name \textit{unplugged}. We further explore the relationship between pre-training and XTRA.

\noindent\textbf{Number of Retrievals}\quad
We experiment with different amounts of retrieved captions during training and inference. In Fig~\ref{fig:ablation}~(a), we show the performance of our method when training with different amounts of retrieval, and different knowledge sources. As can be observed, training with 10 captions and COCO as the knowledge source results in the best performance. In Fig~\ref{fig:ablation}~(b), we show the inference performance for models trained using 10 retrievals. In addition, we show the inference performance of the same model, trained with random amounts of retrieval, between 1 and 20, on the COCO dataset (COCO 20R-C). With this, the best performance is given when we inference with the same amount of trained retrievals, and this then degrades as the number of retrievals differ from how the model was trained. We also see that training with varying number of retrievals achieves better generalization to different amounts of retrievals during inference, as can be seen in Fig~\ref{fig:ablation}~(b), COCO 20R-C, where performance is maintained up to 60 retrievals during inference.

\noindent\textbf{Unplugged Performance}\quad
One interesting observation we make is the ability to "unplug" the knowledge source by not retrieving during inference-time. Tab.~\ref{tab:unplugged} shows a noticeable decrease in performance, indicating the dependency of the reader on the retrieved data during training. When training with COCO as the knowledge source, introducing captions that are very related to the input images is biasing the model to depend on the retrieved captions. For CC and Flickr30K, the domain gap between the downstream task and the knowledge source lessens this gap in unplugged performance. Surprisingly, while ViLBERT performance is generally better than Visual BERT, using our method, the opposite is true when \textit{unplugging} the knowledge source.

\noindent\textbf{External Knowledge Source \& Pre-training}\quad The use of a retrieval mechanism over external knowledge sources raises intriguing questions, e.g.: 1) Is augmentation better than pre-training?; and 2) Can pre-training help the external knowledge source? To address these questions, we experimented with two different pre-training datasets commonly used for VQA - COCO and CC. Tab.~\ref{tab:main_vqa} suggests that for the COCO dataset, using our method is significantly better than pre-training alone, while using pre-training followed by XTRA causes the performance to vary with respect to the reader architecture (e.g., pre-training helps XTRA with ViLBERT, but not with VisualBERT). Tab.~\ref{tab:main_vqa} also shows that fine-tuning our method after pre-training on the same knowledge source yields better performance over pre-training across all knowledge sources and architectures.

\section{Conclusion}
In this work, we presented a novel approach that proposes the use of external knowledge sources in multi-modal prediction models with transformer architectures. We trained a powerful alignment model, DXR, for performing retrieval over external knowledge sources. We showed that our method XTRA yields gains in performance when using an in-domain knowledge source on VQA. We conducted a variety of experiments to show the sensitivity and effects of the knowledge source with various choices for hyperparameters, which shed further light on the different aspects of the model. Future research and applications of our method include interpretability of retrieved data and predictions for verification processes, the demonstration of increased information security by hot-swapping, and unplugged versions of models and new architectures that take advantage of out-of-domain knowledge source. We hope that our approach inspires further work in the direction of hybrid parametric non-parametric models for multi-modal problems.

\bibliography{custom}

\begin{thebibliography}{53}
\expandafter\ifx\csname natexlab\endcsname\relax\def\natexlab#1{#1}\fi

\bibitem[{Chen et~al.(2015)Chen, Fang, Lin, Vedantam, Gupta, Doll{\'a}r, and
  Zitnick}]{chen2015microsoft}
Xinlei Chen, Hao Fang, Tsung-Yi Lin, Ramakrishna Vedantam, Saurabh Gupta, Piotr
  Doll{\'a}r, and C~Lawrence Zitnick. 2015.
\newblock Microsoft coco captions: Data collection and evaluation server.
\newblock \emph{arXiv preprint arXiv:1504.00325}.

\bibitem[{Chen et~al.(2013)Chen, Shrivastava, and Gupta}]{chen2013neil}
Xinlei Chen, Abhinav Shrivastava, and Abhinav Gupta. 2013.
\newblock Neil: Extracting visual knowledge from web data.
\newblock In \emph{Proceedings of the IEEE international conference on computer
  vision}, pages 1409--1416.

\bibitem[{Chen et~al.(2020)Chen, Li, Yu, El~Kholy, Ahmed, Gan, Cheng, and
  Liu}]{chen2020uniter}
Yen-Chun Chen, Linjie Li, Licheng Yu, Ahmed El~Kholy, Faisal Ahmed, Zhe Gan,
  Yu~Cheng, and Jingjing Liu. 2020.
\newblock Uniter: Universal image-text representation learning.
\newblock In \emph{European Conference on Computer Vision}, pages 104--120.
  Springer.

\bibitem[{Devlin et~al.(2018)Devlin, Chang, Lee, and
  Toutanova}]{devlin2018bert}
Jacob Devlin, Ming-Wei Chang, Kenton Lee, and Kristina Toutanova. 2018.
\newblock Bert: Pre-training of deep bidirectional transformers for language
  understanding.
\newblock \emph{arXiv preprint arXiv:1810.04805}.

\bibitem[{Devlin et~al.(2015)Devlin, Gupta, Girshick, Mitchell, and
  Zitnick}]{devlin2015exploring}
Jacob Devlin, Saurabh Gupta, Ross Girshick, Margaret Mitchell, and C~Lawrence
  Zitnick. 2015.
\newblock Exploring nearest neighbor approaches for image captioning.
\newblock \emph{arXiv preprint arXiv:1505.04467}.

\bibitem[{Divvala et~al.(2014)Divvala, Farhadi, and
  Guestrin}]{divvala2014learning}
Santosh~K Divvala, Ali Farhadi, and Carlos Guestrin. 2014.
\newblock Learning everything about anything: Webly-supervised visual concept
  learning.
\newblock In \emph{Proceedings of the IEEE Conference on Computer Vision and
  Pattern Recognition}, pages 3270--3277.

\bibitem[{Faghri et~al.(2017)Faghri, Fleet, Kiros, and
  Fidler}]{faghri2017vse++}
Fartash Faghri, David~J Fleet, Jamie~Ryan Kiros, and Sanja Fidler. 2017.
\newblock Vse++: Improving visual-semantic embeddings with hard negatives.
\newblock \emph{arXiv preprint arXiv:1707.05612}.

\bibitem[{Fan et~al.(2020)Fan, Gardent, Braud, and Bordes}]{fan2020augmenting}
Angela Fan, Claire Gardent, Chloe Braud, and Antoine Bordes. 2020.
\newblock Augmenting transformers with knn-based composite memory for dialogue.
\newblock \emph{arXiv preprint arXiv:2004.12744}.

\bibitem[{Guu et~al.(2020)Guu, Lee, Tung, Pasupat, and Chang}]{guu2020realm}
Kelvin Guu, Kenton Lee, Zora Tung, Panupong Pasupat, and Ming-Wei Chang. 2020.
\newblock Realm: Retrieval-augmented language model pre-training.
\newblock \emph{arXiv preprint arXiv:2002.08909}.

\bibitem[{Huang et~al.(2018)Huang, Wu, Song, and Wang}]{huang2018learning}
Yan Huang, Qi~Wu, Chunfeng Song, and Liang Wang. 2018.
\newblock Learning semantic concepts and order for image and sentence matching.
\newblock In \emph{Proceedings of the IEEE Conference on Computer Vision and
  Pattern Recognition}, pages 6163--6171.

\bibitem[{Izacard and Grave(2020)}]{izacard2020leveraging}
Gautier Izacard and Edouard Grave. 2020.
\newblock Leveraging passage retrieval with generative models for open domain
  question answering.
\newblock \emph{arXiv preprint arXiv:2007.01282}.

\bibitem[{Jia et~al.(2021)Jia, Yang, Xia, Chen, Parekh, Pham, Le, Sung, Li, and
  Duerig}]{jia2021scaling}
Chao Jia, Yinfei Yang, Ye~Xia, Yi-Ting Chen, Zarana Parekh, Hieu Pham, Quoc~V
  Le, Yunhsuan Sung, Zhen Li, and Tom Duerig. 2021.
\newblock Scaling up visual and vision-language representation learning with
  noisy text supervision.
\newblock \emph{arXiv preprint arXiv:2102.05918}.

\bibitem[{Jiang et~al.(2020)Jiang, Misra, Rohrbach, Learned-Miller, and
  Chen}]{jiang2020defense}
Huaizu Jiang, Ishan Misra, Marcus Rohrbach, Erik Learned-Miller, and Xinlei
  Chen. 2020.
\newblock In defense of grid features for visual question answering.
\newblock \emph{arXiv preprint arXiv:2001.03615}.

\bibitem[{Jiang et~al.(2018)Jiang, Natarajan, Chen, Rohrbach, Batra, and
  Parikh}]{jiang2018pythia}
Yu~Jiang, Vivek Natarajan, Xinlei Chen, Marcus Rohrbach, Dhruv Batra, and Devi
  Parikh. 2018.
\newblock Pythia v0. 1: the winning entry to the vqa challenge 2018.
\newblock \emph{arXiv preprint arXiv:1807.09956}.

\bibitem[{Johnson et~al.(2017)Johnson, Douze, and J{\'e}gou}]{JDH17}
Jeff Johnson, Matthijs Douze, and Herv{\'e} J{\'e}gou. 2017.
\newblock Billion-scale similarity search with gpus.
\newblock \emph{arXiv preprint arXiv:1702.08734}.

\bibitem[{Karpathy and Fei-Fei(2015)}]{karpathy2015deep}
Andrej Karpathy and Li~Fei-Fei. 2015.
\newblock Deep visual-semantic alignments for generating image descriptions.
\newblock In \emph{Proceedings of the IEEE conference on computer vision and
  pattern recognition}, pages 3128--3137.

\bibitem[{Karpukhin et~al.(2020)Karpukhin, O{\u{g}}uz, Min, Wu, Edunov, Chen,
  and Yih}]{karpukhin2020dense}
Vladimir Karpukhin, Barlas O{\u{g}}uz, Sewon Min, Ledell Wu, Sergey Edunov,
  Danqi Chen, and Wen-tau Yih. 2020.
\newblock Dense passage retrieval for open-domain question answering.
\newblock \emph{arXiv preprint arXiv:2004.04906}.

\bibitem[{Khandelwal et~al.(2019)Khandelwal, Levy, Jurafsky, Zettlemoyer, and
  Lewis}]{khandelwal2019generalization}
Urvashi Khandelwal, Omer Levy, Dan Jurafsky, Luke Zettlemoyer, and Mike Lewis.
  2019.
\newblock Generalization through memorization: Nearest neighbor language
  models.
\newblock \emph{arXiv preprint arXiv:1911.00172}.

\bibitem[{Kiela et~al.(2019)Kiela, Bhooshan, Firooz, and
  Testuggine}]{kiela2019supervised}
Douwe Kiela, Suvrat Bhooshan, Hamed Firooz, and Davide Testuggine. 2019.
\newblock Supervised multimodal bitransformers for classifying images and text.
\newblock \emph{arXiv preprint arXiv:1909.02950}.

\bibitem[{Krishna et~al.(2017)Krishna, Zhu, Groth, Johnson, Hata, Kravitz,
  Chen, Kalantidis, Li, Shamma et~al.}]{krishna2017visual}
Ranjay Krishna, Yuke Zhu, Oliver Groth, Justin Johnson, Kenji Hata, Joshua
  Kravitz, Stephanie Chen, Yannis Kalantidis, Li-Jia Li, David~A Shamma, et~al.
  2017.
\newblock Visual genome: Connecting language and vision using crowdsourced
  dense image annotations.
\newblock \emph{International journal of computer vision}, 123(1):32--73.

\bibitem[{Lee et~al.(2018)Lee, Chen, Hua, Hu, and He}]{lee2018stacked}
Kuang-Huei Lee, Xi~Chen, Gang Hua, Houdong Hu, and Xiaodong He. 2018.
\newblock Stacked cross attention for image-text matching.
\newblock In \emph{Proceedings of the European Conference on Computer Vision
  (ECCV)}, pages 201--216.

\bibitem[{Lewis et~al.(2020)Lewis, Perez, Piktus, Petroni, Karpukhin, Goyal,
  K{\"u}ttler, Lewis, Yih, Rockt{\"a}schel et~al.}]{lewis2020retrieval}
Patrick Lewis, Ethan Perez, Aleksandara Piktus, Fabio Petroni, Vladimir
  Karpukhin, Naman Goyal, Heinrich K{\"u}ttler, Mike Lewis, Wen-tau Yih, Tim
  Rockt{\"a}schel, et~al. 2020.
\newblock Retrieval-augmented generation for knowledge-intensive nlp tasks.
\newblock \emph{arXiv preprint arXiv:2005.11401}.

\bibitem[{Li et~al.(2020{\natexlab{a}})Li, Duan, Fang, Gong, Jiang, and
  Zhou}]{li2020unicoder}
Gen Li, Nan Duan, Yuejian Fang, Ming Gong, Daxin Jiang, and Ming Zhou.
  2020{\natexlab{a}}.
\newblock Unicoder-vl: A universal encoder for vision and language by
  cross-modal pre-training.
\newblock In \emph{AAAI}, pages 11336--11344.

\bibitem[{Li et~al.(2019{\natexlab{a}})Li, Zhang, Li, Li, and
  Fu}]{li2019visual}
Kunpeng Li, Yulun Zhang, Kai Li, Yuanyuan Li, and Yun Fu. 2019{\natexlab{a}}.
\newblock Visual semantic reasoning for image-text matching.
\newblock In \emph{Proceedings of the IEEE International Conference on Computer
  Vision}, pages 4654--4662.

\bibitem[{Li et~al.(2019{\natexlab{b}})Li, Yatskar, Yin, Hsieh, and
  Chang}]{li2019visualbert}
Liunian~Harold Li, Mark Yatskar, Da~Yin, Cho-Jui Hsieh, and Kai-Wei Chang.
  2019{\natexlab{b}}.
\newblock Visualbert: A simple and performant baseline for vision and language.
\newblock \emph{arXiv preprint arXiv:1908.03557}.

\bibitem[{Li et~al.(2020{\natexlab{b}})Li, Yin, Li, Zhang, Hu, Zhang, Wang, Hu,
  Dong, Wei et~al.}]{li2020oscar}
Xiujun Li, Xi~Yin, Chunyuan Li, Pengchuan Zhang, Xiaowei Hu, Lei Zhang, Lijuan
  Wang, Houdong Hu, Li~Dong, Furu Wei, et~al. 2020{\natexlab{b}}.
\newblock Oscar: Object-semantics aligned pre-training for vision-language
  tasks.
\newblock In \emph{European Conference on Computer Vision}, pages 121--137.
  Springer.

\bibitem[{Liu et~al.(2017)Liu, Guo, Bakker, and Lew}]{liu2017learning}
Yu~Liu, Yanming Guo, Erwin~M Bakker, and Michael~S Lew. 2017.
\newblock Learning a recurrent residual fusion network for multimodal matching.
\newblock In \emph{Proceedings of the IEEE International Conference on Computer
  Vision}, pages 4107--4116.

\bibitem[{Lu et~al.(2019)Lu, Batra, Parikh, and Lee}]{lu2019vilbert}
Jiasen Lu, Dhruv Batra, Devi Parikh, and Stefan Lee. 2019.
\newblock Vilbert: Pretraining task-agnostic visiolinguistic representations
  for vision-and-language tasks.
\newblock In \emph{Advances in Neural Information Processing Systems}, pages
  13--23.

\bibitem[{Lu et~al.(2021)Lu, Zhao, and Lee}]{lu2021visualsparta}
Xiaopeng Lu, Tiancheng Zhao, and Kyusong Lee. 2021.
\newblock Visualsparta: Sparse transformer fragment-level matching for
  large-scale text-to-image search.
\newblock \emph{arXiv preprint arXiv:2101.00265}.

\bibitem[{Marino et~al.(2019)Marino, Rastegari, Farhadi, and
  Mottaghi}]{marino2019ok}
Kenneth Marino, Mohammad Rastegari, Ali Farhadi, and Roozbeh Mottaghi. 2019.
\newblock Ok-vqa: A visual question answering benchmark requiring external
  knowledge.
\newblock In \emph{Proceedings of the IEEE Conference on Computer Vision and
  Pattern Recognition}, pages 3195--3204.

\bibitem[{Messina et~al.(2020{\natexlab{a}})Messina, Amato, Esuli, Falchi,
  Gennaro, and Marchand-Maillet}]{messina2020fine}
Nicola Messina, Giuseppe Amato, Andrea Esuli, Fabrizio Falchi, Claudio Gennaro,
  and St{\'e}phane Marchand-Maillet. 2020{\natexlab{a}}.
\newblock Fine-grained visual textual alignment for cross-modal retrieval using
  transformer encoders.
\newblock \emph{arXiv preprint arXiv:2008.05231}.

\bibitem[{Messina et~al.(2020{\natexlab{b}})Messina, Falchi, Esuli, and
  Amato}]{messina2020transformer}
Nicola Messina, Fabrizio Falchi, Andrea Esuli, and Giuseppe Amato.
  2020{\natexlab{b}}.
\newblock Transformer reasoning network for image-text matching and retrieval.
\newblock \emph{arXiv preprint arXiv:2004.09144}.

\bibitem[{Narasimhan et~al.(2018)Narasimhan, Lazebnik, and
  Schwing}]{narasimhan2018out}
Medhini Narasimhan, Svetlana Lazebnik, and Alexander Schwing. 2018.
\newblock Out of the box: Reasoning with graph convolution nets for factual
  visual question answering.
\newblock In \emph{Advances in neural information processing systems}, pages
  2654--2665.

\bibitem[{Narasimhan and Schwing(2018)}]{narasimhan2018straight}
Medhini Narasimhan and Alexander~G Schwing. 2018.
\newblock Straight to the facts: Learning knowledge base retrieval for factual
  visual question answering.
\newblock In \emph{Proceedings of the European conference on computer vision
  (ECCV)}, pages 451--468.

\bibitem[{Nguyen et~al.(2020)Nguyen, Goswami, and Chen}]{nguyen2020revisiting}
Duy-Kien Nguyen, Vedanuj Goswami, and Xinlei Chen. 2020.
\newblock Revisiting modulated convolutions for visual counting and beyond.
\newblock \emph{arXiv preprint arXiv:2004.11883}.

\bibitem[{Radford et~al.(2021)Radford, Kim, Hallacy, Ramesh, Goh, Agarwal,
  Sastry, Askell, Mishkin, Clark et~al.}]{radford2021learning}
Alec Radford, Jong~Wook Kim, Chris Hallacy, Aditya Ramesh, Gabriel Goh,
  Sandhini Agarwal, Girish Sastry, Amanda Askell, Pamela Mishkin, Jack Clark,
  et~al. 2021.
\newblock Learning transferable visual models from natural language
  supervision.
\newblock \emph{arXiv preprint arXiv:2103.00020}.

\bibitem[{Sadeghi et~al.(2015)Sadeghi, Kumar~Divvala, and
  Farhadi}]{sadeghi2015viske}
Fereshteh Sadeghi, Santosh~K Kumar~Divvala, and Ali Farhadi. 2015.
\newblock Viske: Visual knowledge extraction and question answering by visual
  verification of relation phrases.
\newblock In \emph{Proceedings of the IEEE conference on computer vision and
  pattern recognition}, pages 1456--1464.

\bibitem[{Sharma et~al.(2018)Sharma, Ding, Goodman, and
  Soricut}]{sharma2018conceptual}
Piyush Sharma, Nan Ding, Sebastian Goodman, and Radu Soricut. 2018.
\newblock Conceptual captions: A cleaned, hypernymed, image alt-text dataset
  for automatic image captioning.
\newblock In \emph{Proceedings of the 56th Annual Meeting of the Association
  for Computational Linguistics (Volume 1: Long Papers)}, pages 2556--2565.

\bibitem[{Singh et~al.(2020{\natexlab{a}})Singh, Goswami, Natarajan, Jiang,
  Chen, Shah, Rohrbach, Batra, and Parikh}]{singh2020mmf}
Amanpreet Singh, Vedanuj Goswami, Vivek Natarajan, Yu~Jiang, Xinlei Chen, Meet
  Shah, Marcus Rohrbach, Dhruv Batra, and Devi Parikh. 2020{\natexlab{a}}.
\newblock Mmf: A multimodal framework for vision and language research.
\newblock \url{https://github.com/facebookresearch/mmf}.

\bibitem[{Singh et~al.(2020{\natexlab{b}})Singh, Goswami, and
  Parikh}]{singh2020we}
Amanpreet Singh, Vedanuj Goswami, and Devi Parikh. 2020{\natexlab{b}}.
\newblock Are we pretraining it right? digging deeper into visio-linguistic
  pretraining.
\newblock \emph{arXiv preprint arXiv:2004.08744}.

\bibitem[{Su et~al.(2019)Su, Zhu, Cao, Li, Lu, Wei, and Dai}]{su2019vl}
Weijie Su, Xizhou Zhu, Yue Cao, Bin Li, Lewei Lu, Furu Wei, and Jifeng Dai.
  2019.
\newblock Vl-bert: Pre-training of generic visual-linguistic representations.
\newblock \emph{arXiv preprint arXiv:1908.08530}.

\bibitem[{Verga et~al.(2020)Verga, Sun, Soares, and Cohen}]{verga2020facts}
Pat Verga, Haitian Sun, Livio~Baldini Soares, and William~W Cohen. 2020.
\newblock Facts as experts: Adaptable and interpretable neural memory over
  symbolic knowledge.
\newblock \emph{arXiv preprint arXiv:2007.00849}.

\bibitem[{Wang et~al.(2018)Wang, Wu, Shen, Dick, and van~den
  Hengel}]{wang2018fvqa}
Peng Wang, Qi~Wu, Chunhua Shen, Anthony Dick, and Anton van~den Hengel. 2018.
\newblock Fvqa: Fact-based visual question answering.
\newblock \emph{IEEE transactions on pattern analysis and machine
  intelligence}, 40(10):2413--2427.

\bibitem[{Wang et~al.(2015)Wang, Wu, Shen, Hengel, and Dick}]{wang2015explicit}
Peng Wang, Qi~Wu, Chunhua Shen, Anton van~den Hengel, and Anthony Dick. 2015.
\newblock Explicit knowledge-based reasoning for visual question answering.
\newblock \emph{arXiv preprint arXiv:1511.02570}.

\bibitem[{Wang et~al.(2019)Wang, Liu, Li, Sheng, Yan, Wang, and
  Shao}]{wang2019camp}
Zihao Wang, Xihui Liu, Hongsheng Li, Lu~Sheng, Junjie Yan, Xiaogang Wang, and
  Jing Shao. 2019.
\newblock Camp: Cross-modal adaptive message passing for text-image retrieval.
\newblock In \emph{Proceedings of the IEEE International Conference on Computer
  Vision}, pages 5764--5773.

\bibitem[{Wei et~al.(2020)Wei, Zhang, Li, Zhang, and Wu}]{wei2020multi}
Xi~Wei, Tianzhu Zhang, Yan Li, Yongdong Zhang, and Feng Wu. 2020.
\newblock Multi-modality cross attention network for image and sentence
  matching.
\newblock In \emph{Proceedings of the IEEE/CVF Conference on Computer Vision
  and Pattern Recognition}, pages 10941--10950.

\bibitem[{Wu et~al.(2019)Wu, Wang, Song, and Huang}]{wu2019learning}
Yiling Wu, Shuhui Wang, Guoli Song, and Qingming Huang. 2019.
\newblock Learning fragment self-attention embeddings for image-text matching.
\newblock In \emph{Proceedings of the 27th ACM International Conference on
  Multimedia}, pages 2088--2096.

\bibitem[{Young et~al.(2014)Young, Lai, Hodosh, and
  Hockenmaier}]{young2014image}
Peter Young, Alice Lai, Micah Hodosh, and Julia Hockenmaier. 2014.
\newblock From image descriptions to visual denotations: New similarity metrics
  for semantic inference over event descriptions.
\newblock \emph{Transactions of the Association for Computational Linguistics},
  2:67--78.

\bibitem[{Zhang and Lu(2018)}]{zhang2018deep}
Ying Zhang and Huchuan Lu. 2018.
\newblock Deep cross-modal projection learning for image-text matching.
\newblock In \emph{Proceedings of the European Conference on Computer Vision
  (ECCV)}, pages 686--701.

\bibitem[{Zheng et~al.(2017)Zheng, Zheng, Garrett, Yang, and
  Shen}]{zheng2017dual}
Zhedong Zheng, Liang Zheng, Michael Garrett, Yi~Yang, and Yi-Dong Shen. 2017.
\newblock Dual-path convolutional image-text embedding. corr abs/1711.05535
  (2017).
\newblock \emph{arXiv preprint arXiv:1711.05535}.

\bibitem[{Zhu et~al.(2014)Zhu, Fathi, and Fei-Fei}]{zhu2014reasoning}
Yuke Zhu, Alireza Fathi, and Li~Fei-Fei. 2014.
\newblock Reasoning about object affordances in a knowledge base
  representation.
\newblock In \emph{European conference on computer vision}, pages 408--424.
  Springer.

\bibitem[{Zhu et~al.(2017)Zhu, Lim, and Fei-Fei}]{zhu2017knowledge}
Yuke Zhu, Joseph~J Lim, and Li~Fei-Fei. 2017.
\newblock Knowledge acquisition for visual question answering via iterative
  querying.
\newblock In \emph{Proceedings of the IEEE Conference on Computer Vision and
  Pattern Recognition}, pages 1154--1163.

\bibitem[{Zhu et~al.(2015)Zhu, Zhang, R{\'e}, and Fei-Fei}]{zhu2015building}
Yuke Zhu, Ce~Zhang, Christopher R{\'e}, and Li~Fei-Fei. 2015.
\newblock Building a large-scale multimodal knowledge base system for answering
  visual queries.
\newblock \emph{arXiv preprint arXiv:1507.05670}.

\end{thebibliography}
\bibliographystyle{acl_natbib}

\clearpage
\appendix
\section{Retrieval}
Tab.~\ref{tab:alignmet_full} shows a complete comparison of the different alignment methods in the cross-modal alignment literature. The top part corresponds to methods which use vector representations, grid-features, and do not share information between the modality branches. The bottom part shows the rest of the methods.
\label{ap:retrieval}
\begin{table*}[t]
    \centering
    \begin{tabular*}{\linewidth}{@{\extracolsep{\fill}}lcccccc}
        \toprule
         & \multicolumn{3}{c}{Text $\rightarrow$ Image} & \multicolumn{3}{c}{Image $\rightarrow$ Text}\\
        Method & R@1 & R@5 & R@10 & R@1 & R@5 & R@10\\
        \midrule
        RRF & 35.4 & 68.3 & 79.9 & 47.6 & 77.4 & 87.1\\ 
        CMPM & 37.3 & 65.7 & 75.5 & 49.6 & 76.8 & 86.1\\
        DPC & 39.1 & 69.2 & 69.2 & 55.6 & 81.9 & 89.5\\
        VSE++ & 39.6 & 69.6 & 79.5 & 52.9 & 79.1 & 87.2\\
        \textbf{DXR} & \textbf{50.6} & \textbf{78.8} & \textbf{86.7} & \textbf{65.1} & \textbf{87.3} & \textbf{92.6}\\
        \midrule
        CLIP$^\dagger$ & 68.7 & 90.6 & 95.2 & 88.0 & 98.7 & 99.4 \\
        ALIGN$^\dagger$ & 75.7 & 93.8 & 96.8 & 88.6 & 98.7 & 99.7 \\
        \midrule
        TERN & 41.1 & 71.9 & 81.2 & 53.2 & 79.4 & 86.0 \\
        SCO & 41.1 & 70.5 & 80.1 & 55.5 & 82.0 & 89.3\\
        SAEM & 52.4 & 81.1 & 88.1 & 69.1 & 91.0 & 95.1 \\
        SCAN & 48.6 & 77.7 & 85.2 & 67.4 & 90.3 & 95.8 \\
        CAMP & 51.5 & 77.1 & 85.3 & 68.1 & 89.7 & 95.2\\
        VSRN & 54.7 & 81.8 & 88.2 & 71.3 & 90.6 & 96.0\\
        TERAN & 56.5 & 81.2 & 88.2 & 70.8 & 90.9 & 95.5\\
        MMCA & 54.8 & 81.4 & 87.8 & 74.2 & 92.8 & 96.4\\
        Unicoder-VL &  71.5 & 90.9 & 94.9 &  86.2 & 96.3 & 99.0\\
        UNITER & 73.6 & 93.0 & 95.9 & 88.2 & 98.4 & 99.0\\
        \bottomrule
    \end{tabular*}
    \caption{Retrieval results for Flickr30K. \textbf{Top} - methods that use raw images as input, and vector representations for the embedding space. \textbf{Bottom} Methods that use detection features or sequence similarity measures. We denote by $\dagger$ methods that train on substantial amount of novel data.}
    \label{tab:ret_flickr_full}
\end{table*}
\begin{table*}[t]
    \centering
    \begin{tabular*}{\linewidth}{@{\extracolsep{\fill}}lc@{~~}c@{~~}cc@{~~}c@{~~}c|c@{~~}c@{~~}cc@{~~}c@{~~}c}
        \toprule
         & \multicolumn{6}{c}{COCO 1K} & \multicolumn{6}{c}{COCO 5K}\\
         & \multicolumn{3}{c}{Text $\rightarrow$ Image} & \multicolumn{3}{c}{Image $\rightarrow$ Text} & \multicolumn{3}{c}{Text $\rightarrow$ Image} & \multicolumn{3}{c}{Image $\rightarrow$ Text}\\
        Method & R@1 & R@5 & R@10 & R@1 & R@5 & R@10 & R@1 & R@5 & R@10 & R@1 & R@5 & R@10\\
        \midrule
        DPC & 47.1 & 79.9 & 90.0 & 65.6 & 89.8 & 95.5 & 25.3 & 53.4 & 66.4 & 41.2 & 70.5 & 81.1\\
        VSE++ & 52.0 & 83.1 & 92.0 & 64.6 & 89.1 & 95.7 & 30.3 & 59.1 & 72.4 & 41.3 & 69.2 & 81.2\\
        CMPM & 44.6 & 78.8 & 89.0 & 56.1 & 86.3 & 92.9 & 22.9 & 50.2 & 63.8 & 31.1 & 60.7 & 73.9\\
        \textbf{DXR} & \textbf{56.8} & \textbf{88.2} & \textbf{94.9} & \textbf{67.0} & \textbf{93.0} & \textbf{97.6} & \textbf{33.9} & \textbf{64.9} & \textbf{77.4} & \textbf{44.9} & \textbf{75.2} & \textbf{84.7}\\
        \midrule
        CLIP$^\dagger$ & - & - & - & - & - & - & 37.8 & 62.4 & 72.2 & 58.4 & 81.5 & 88.1\\
        ALIGN$^\dagger$ & - & - & - & - & - & - & 45.6 & 69.8 & 78.6 & 58.6 & 83.0 & 89.7\\
        \midrule
        TERN & 51.9 & 85.6 & 93.6 & 63.7 & 90.5 & 96.2 & 28.7 & 59.7 & 72.7 & 38.4 & 69.5 & 81.3 \\
        SCO & 56.7 & 87.5 & 94.8 & 69.9 & 92.9 & 97.5 & 33.1 & 62.9 & 75.5 & 42.8 & 72.3 & 83.0\\
        SAEM & 57.8 & 88.6 & 94.9 & 71.2 & 94.1 & 97.7 & - & - & - & - & - & - \\
        SCAN & 58.8 & 88.4 & 94.8 & 72.7 & 94.8 & 98.4    & 38.6 & 69.3 & 80.4 & 50.4 & 82.2 & 90.0 \\
        CAMP & 58.5 & 87.9 & 95.0 & 72.3 & 94.8 & 98.3 & 39.0 & 68.9 & 80.2 & 50.1 & 82.1 & 89.7\\
        VSRN & 62.8 & 89.7 & 95.1 & 76.2 & 94.8 & 98.2 & 40.5 & 70.6 & 81.1 & 53.0 & 81.1 & 89.4\\
        TERAN & 65.0 & 91.2 & 96.4 & 77.7 & 95.9 & 98.6 & 42.6 & 72.5 & 82.9 & 55.6 & 83.9 & 91.6\\
        MMCA & 61.6 & 89.8 & 95.2 & 74.8 & 95.6 & 97.7 &  38.7 & 69.7 & 80.8 & 54.0 & 82.5 & 90.7\\
        Unicoder-VL &  69.7 & 93.5 & 97.2 & 84.3 & 97.3 & 99.3 & 46.7 & 76.0 & 85.3 & 62.3 & 87.1 & 92.8\\
        UNITER & - & - & - & - & - & - & 51.7 & 78.4 & 86.9 & 66.6 & 89.4 & 94.2\\
        Oscar & 78.2 & 95.8 & 98.3 & 89.8 & 98.8 & 99.7 & 57.5 & 82.8 & 89.8 & 73.5 & 92.2 & 96.0\\
        \bottomrule
    \end{tabular*}
    \caption{Retrieval results for COCO. \textbf{Top} - methods that use raw images as input, and vector representations for the embedding space. \textbf{Bottom} Methods that use detection features or sequence similarity measures. We denote by $\dagger$ methods that train on substantial amount of novel data.}
    \label{tab:alignmet_full}
\end{table*}

\end{document}